\documentclass[preprint,12pt]{elsarticle}
\usepackage[utf8]{inputenc}
\usepackage[russian,english]{babel}
\usepackage{amsfonts}
\usepackage{graphicx}
\usepackage{ulem}

\journal{Information Sciences}

\begin{document}

\begin{frontmatter}

\title{In narrative texts punctuation marks obey the same statistics as words}

\author{Andrzej Kulig$^{1}$, Jaros{\l}aw Kwapie\'n$^{1}$, Tomasz Stanisz$^{1}$, Stanis\l{}aw Dro\.zd\.z$^{1,2}$\corref{ref::corrauth}}
\cortext[ref::corrauth]{Corresponding author: stanislaw.drozdz@ifj.edu.pl}
\address{$^{1}$ Complex Systems Theory Department, Institute of Nuclear Physics, Polish Academy of Sciences, ul.~Radzikowskiego 152, 31-342 Krak\'ow, Poland}
\address{$^{2}$ Faculty of Physics, Mathematics and Computer Science, Cracow University of Technology, ul.~Warszawska 24, 31-155 Krak\'ow, Poland}

\begin{abstract}

From a grammar point of view, the role of punctuation marks in a sentence is formally defined and well understood. In semantic analysis punctuation plays also a crucial role as a method of avoiding ambiguity of the meaning. A different situation can be observed in the statistical analyses of language samples, where the decision on whether the punctuation marks should be considered or should be neglected is seen rather as arbitrary and at present it belongs to a researcher's preference. An objective of this work is to shed some light onto this problem by providing us with an answer to the question whether the punctuation marks may be treated as ordinary words and whether they should be included in any analysis of the word co-occurences. We already know from our previous study (S.~Dro\.zd\.z {\it et al.}, Inf. Sci. 331 (2016) 32-44) that full stops that determine the length of sentences are the main carrier of long-range correlations. Now we extend that study and analyze statistical properties of the most common punctuation marks in a few Indo-European languages, investigate their frequencies, and locate them accordingly in the Zipf rank-frequency plots as well as study their role in the word-adjacency networks. We show that, from a statistical viewpoint, the punctuation marks reveal properties that are qualitatively similar to the properties of the most frequent words like articles, conjunctions, pronouns, and prepositions. This refers to both the Zipfian analysis and the network analysis. By adding the punctuation marks to the Zipf plots, we also show that these plots that are normally described by the Zipf-Mandelbrot distribution largely restore the power-law Zipfian behaviour for the most frequent items.

Our results indicate that the punctuation marks can fruitfully be considered in the linguistic studies as their inclusion effectively extends dimensionality of an analysis and, therefore, it opens more space for possible manifestation of some previously unobserved effects.

\end{abstract}


\end{frontmatter}

\section{Introduction}

Natural language is one of the most vivid examples of complex systems~\cite{kwapien2012}, where the term \textit{more is different}~\cite{anderson1972} like no other succinctly defines its features. Indeed, the relatively small number of elementary items, the phonemes and letters, allow one to create more complex elements: the words. They form references to everything that a human can name and describe. However, the words alone do not constitute the whole essence of language and another complex entity is a prerequisite here: the sentence~\cite{drozdz2016}. The sentential structure is a standard feature of almost all written languages. Only at this level the semantics in its whole richness and with a variety of carriers emerges: words, syntax, phrases, clauses, and punctuation in written language.

Statistical analyses of language samples that were carried out since over a century ago~\cite{estoup1916,zipf1932} revealed the existence of laws that describe language quantitatively. Classical statistical study comprises, among others, the empirical word frequency distribution that is compared with the power-law model known as the Zipf law~\cite{zipf1949} or its generalized form known as the Zipf-Mandelbrot law~\cite{mandelbrot1965,montemurro2001}, and the functional relation between the length of a text and the number of unique words used to compose it, modelled by the Heaps law~\cite{herdan1960,heaps1978,gerlach2013}. A relatively new approach is a description of language in the network formalism~\cite{ferrer2001,dorogovtsev2001,masucci2006,markosova2008,grabska2012} that, among others, reveals that certain network representations of the lexical structure of texts (e.g. the word co-occurrence) belong to the scale-free class, similar to the semantic networks constructed based on the meaning of words~\cite{liu2009,amancio2012,amancio2015}.

Writing requires the use of punctuation; otherwise some expressions might be ambiguous and deceptive. Punctuation also allows one to denote separate logical units into which any compound message can be divided. From this perspective, the punctuation marks are something more than merely technical signs serving to allow a reader to comprehend the consecutive pieces of texts more easily. If put in between the words, they also acquire meaning and become meaningful not less than, for example, some words playing mainly grammatical role as conjunctions and articles. For example, even though the full stops do not have clear phonetic expression, they define the length of sentences and thus they can influence a reader's subjective perception of the message content: the speed of events, the descriptive complexity of a given situation, etc. Our recent study shows additionally that punctuation carries long-range correlations in narrative texts~\cite{drozdz2016}. This brings us more quantifiable evidence that punctuation, even though ``silent'', is no less important than words.

Thus, it might seem intuitively natural to include such marks in any analysis, in which the ordinary words are considered: the rank-frequency, the word co-occurrence, and other types of the statistical analyses~\cite{ausloos2010}. It is sometimes done so in the engineering sciences like natural language processing due to practical reasons~\cite{kao2007}, but without any deeper linguistic justification. On the other hand, such an inclusion might not be recommended if the statistical properties of the punctuation marks were significantly different from the corresponding properties of the ordinary words as it would actually mean that the punctuation marks were something different than words. So, this issue appears to be rather a complex one. In order to resolve it, in this work we study the rank-frequency distributions and the word-adjacency networks in the corpora, in which the punctuation marks are treated as words, and compare the results for the punctuation marks with the results for the ordinary words. We argue that these results, which are complementary to the earlier ones published in~\cite{drozdz2016}, can provide one with indication on how to improve reliability of the statistical calculations based on large corpora of the written language samples.

\section{Data and methods}

A literary form that is relatively the closest to the spoken language - prose - is expected to reflect the statistical properties of language. In order to analyze it, we selected a set of well-known novels written in one of six Indo-European languages belonging to the Germanic (English and German), Romance (French and Italian), and Slavic (Polish and Russian) language groups. Our selection criterion was the substantial length of each text sample, i.e., at least 5,000 sentences, which we have already veryfied to be sufficient for a statistical analysis~\cite{drozdz2016}. The texts were downloaded from the Project Gutenberg website~\cite{gutenberg.org}. Apart from the individual texts, we also created 6 monolingual corpora by merging together at least 5 texts written in the same language so that each corpus consisted of about one million words $-$ a volume that was sufficient for our statistical analysis (see Appendix for a list of texts).

Some redundant words residing outside the sentence structure of texts (such as {\it chapter}, {\it part}, {\it epilogue}, etc.), footnotes, page numbers, and typographic marks (quotation marks, parentheses, etc.) were deleted. All standard abbreviations specific to a given language (like {\it Mrs.} and {\it Dr.} in English) were cleaned of dots and counted as separate words. The following marks were considered the full stops that end a sentence: dots, question marks, exclamation marks, and ellipses. Apart from the full stops, our analysis also included commas, colons, and semicolons.

Moreover, the notion of the punctuation marks may be generalized in such a way that it includes new chapters, new parts, and new paragraphs (that are recognized as the separators stronger than a full stop), as well as new lines (that may further be divided into: comma-new line, colon-new line, etc.). While the division into parts is too sparse to be meaningful in our analysis and the localization of all new paragraphs and new lines is too demanding to be easily done here, we extended our analysis over the chapters. In each text we found the places, in which new chapters begin, and introduced them into the texts as an additional punctuation mark (denoted as \#chap). We prefered not to consider any specific word as a separator in this context, because different ways of denoting new chapters are used in different texts: the word ``chapter'', the Roman or the Hindu-Arabic numerals, the asterisks, or even just the voids. One issue should be kept in mind, however. While the standard punctuation can be viewed as an inherent part of the natural language that helps one to understand the message, the division of texts into paragraphs, chapters, and parts is purely a writing technique not necessary from the point of view of the language organization.

Our first analysis was based on the frequency of word occurrence in a sample, which is a standard approach. It allowed us to check for possible statistical similarities between the punctuation marks and the ordinary words. It also aimed at testing whether these additional elements obey the well-known empirical Zipf law. Next, in a word-adjacency network representation, where nodes represent words and connections represent the words' adjacent positions, the punctuation marks were taken into account like usual words. Doing so has practical importance for the consistency of the network creation process: otherwise there might be a problem whether the node representing a word ending a sentence and the node representing a word that starts the subsequent sentence may be connected to each other. On the one hand, such words are more loosely related semantically than the words within the same sentence are, but, on the other hand, leaving those nodes unconnected can lead to the formation of a disconnected network, for which many useful network measures cannot be well-defined. Identification of the punctuation marks as words thus allowed us to overcome this difficulty and to apply all the standard network measures effectively. 

All calculations were performed in Mathematica and C++ environments independently. For better comparison between the corresponding results, all respective figures are shown in the same scale ranges.

\section{Main results}

\subsection{Zipf analyses for language samples}

The primary characteristics of natural language samples describing its quantitative structure is the Zipf distribution. It states that the probability $P(R)$ of encountering the $R$th most frequent word scales according to $P(R) \sim R^{-\alpha}$ for $\alpha \approx 1$. The Zipfian scaling in its original formulation holds for the majority of ranks except for a few highest ones, where the power law breaks and the correspodning plots are deflected towards lower frequencies than those expected from the pure power law. Therefore a better agreement with the empirical data one can obtain using the so-called Zipf-Mandelbrot law (shifted power-law):
\begin{equation}
P(R) \sim (R+c)^{-\alpha},
\label{eq::zipf.mandelbrot}
\end{equation}
where $c$ is the parameter responsible for the above-mentioned deflection. There are different hypotheses on the origin of the Zipf law, with the principle of least effort~\cite{zipf1949} and the communication optimization~\cite{mandelbrot1953} among them. It should be noted that this situation occurs only if a language sample is created in the unconstrained and spontaneous conditions. Existing aberrations from a power-law regime have appropriate justifications that have their source in an intellectual disability~\cite{piotrowska2004} or in sophisticated creative workshops~\cite{kwapien2010}.

\begin{figure}[!h]
\centering
\includegraphics[scale=0.44]{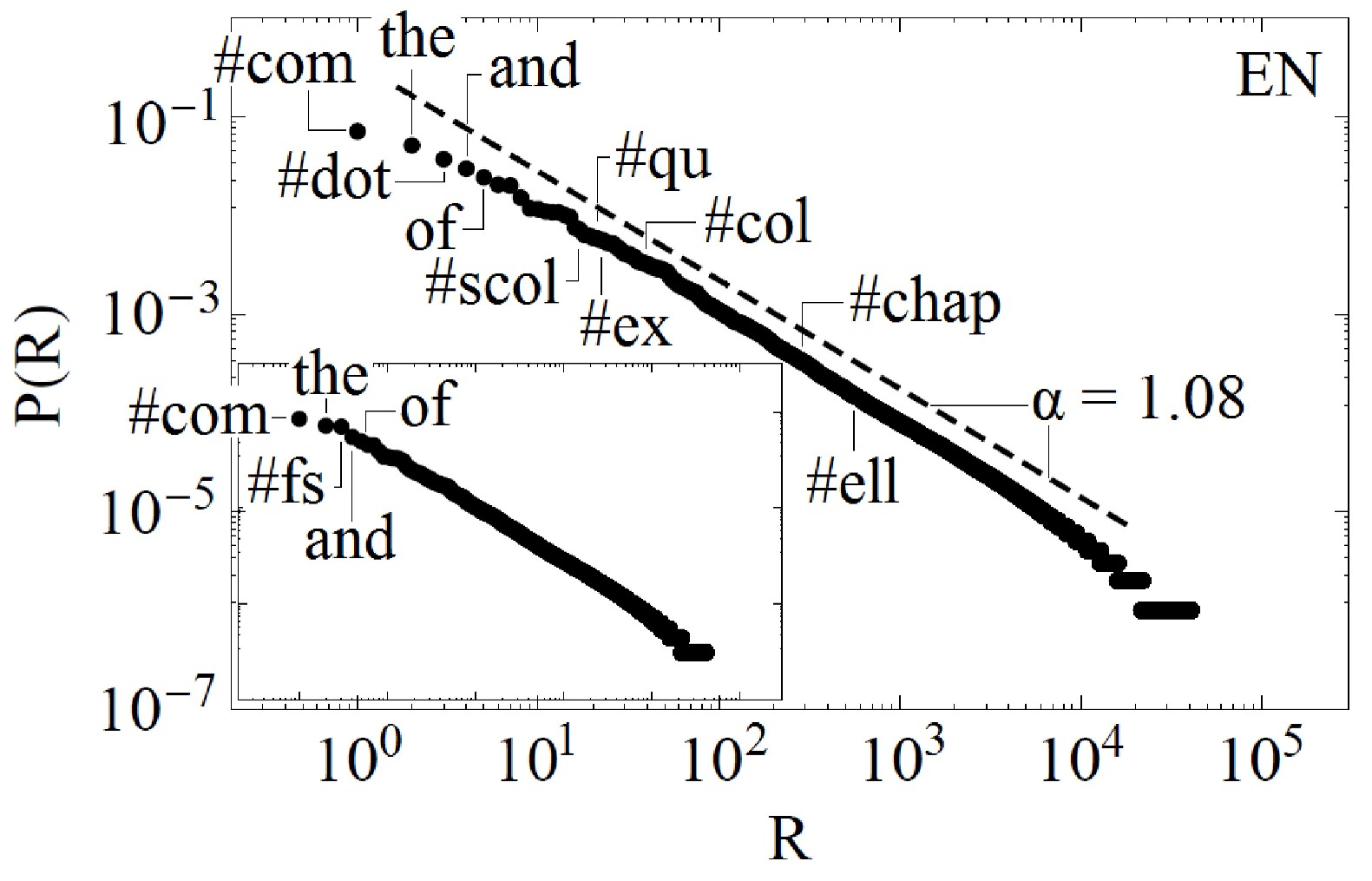}
\includegraphics[scale=0.44]{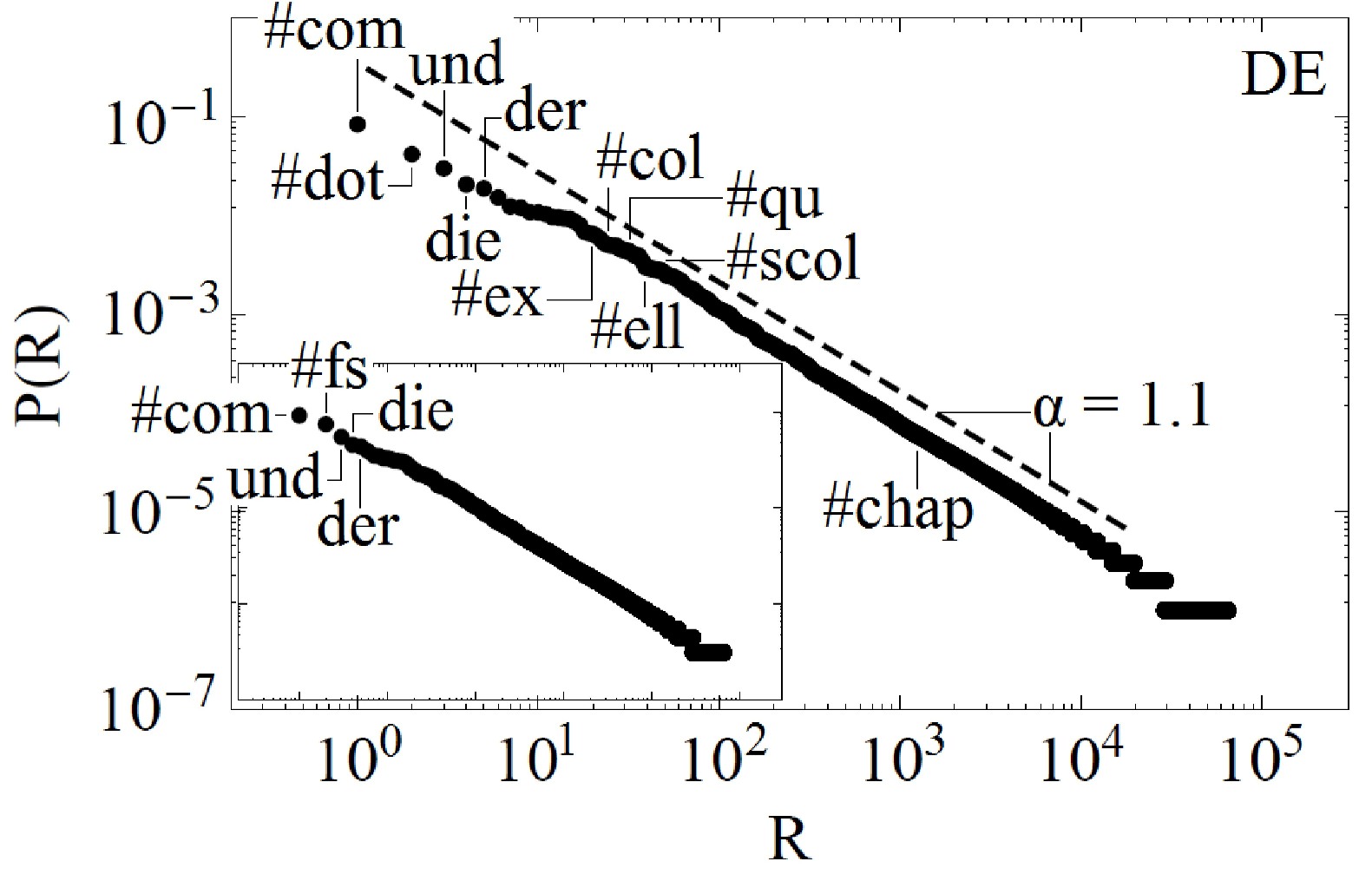}

\includegraphics[scale=0.44]{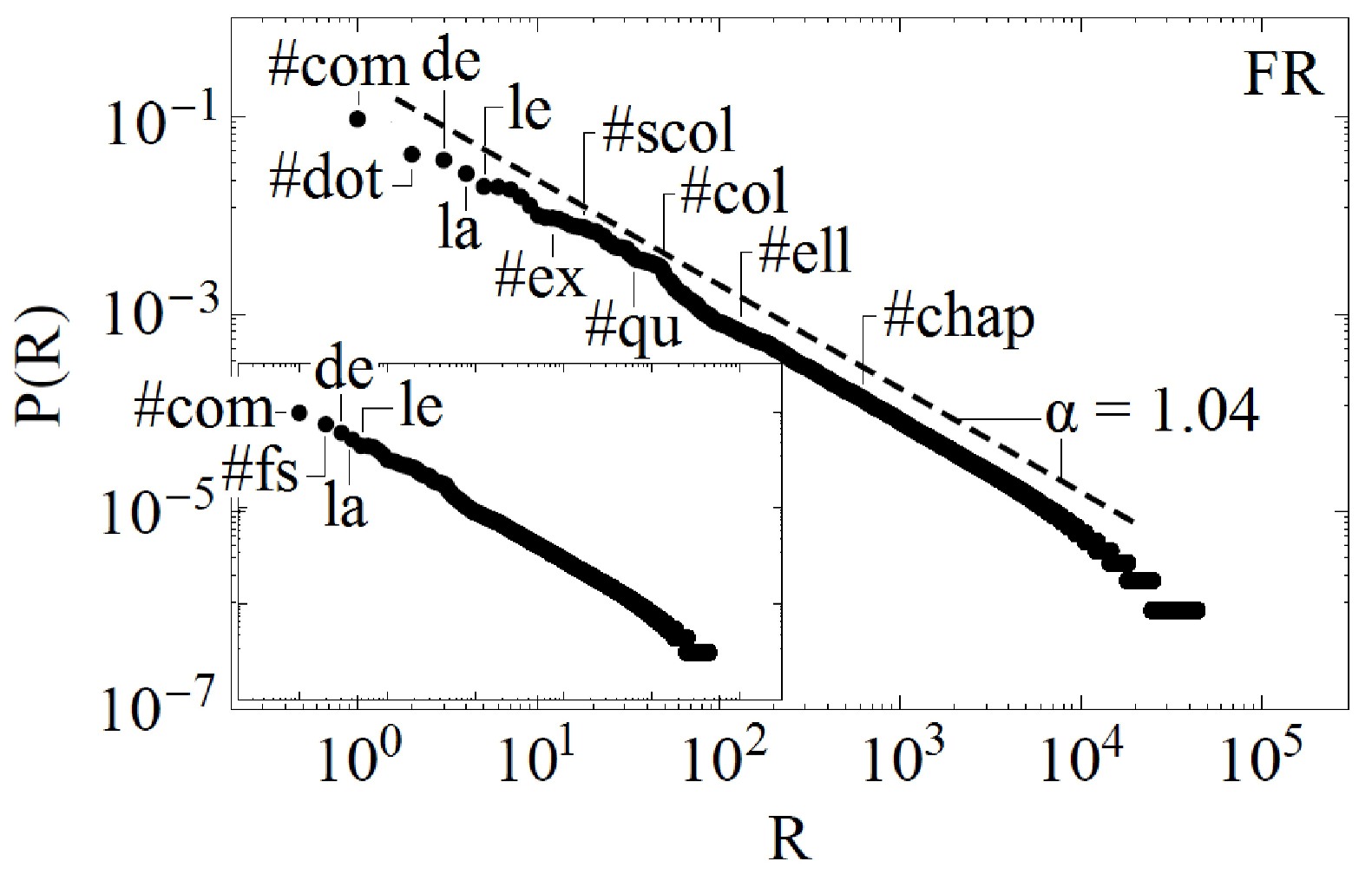}
\includegraphics[scale=0.44]{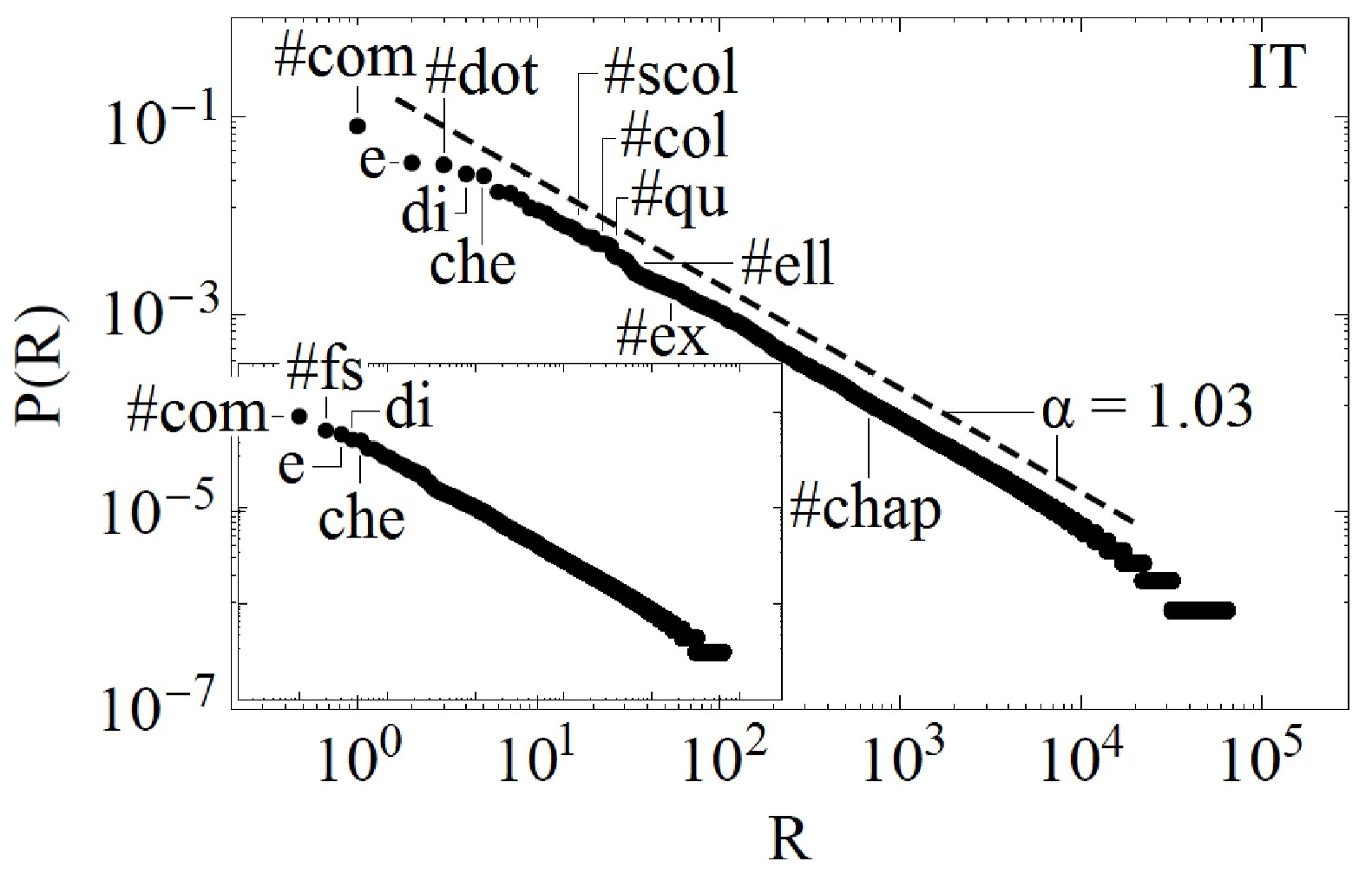}

\includegraphics[scale=0.44]{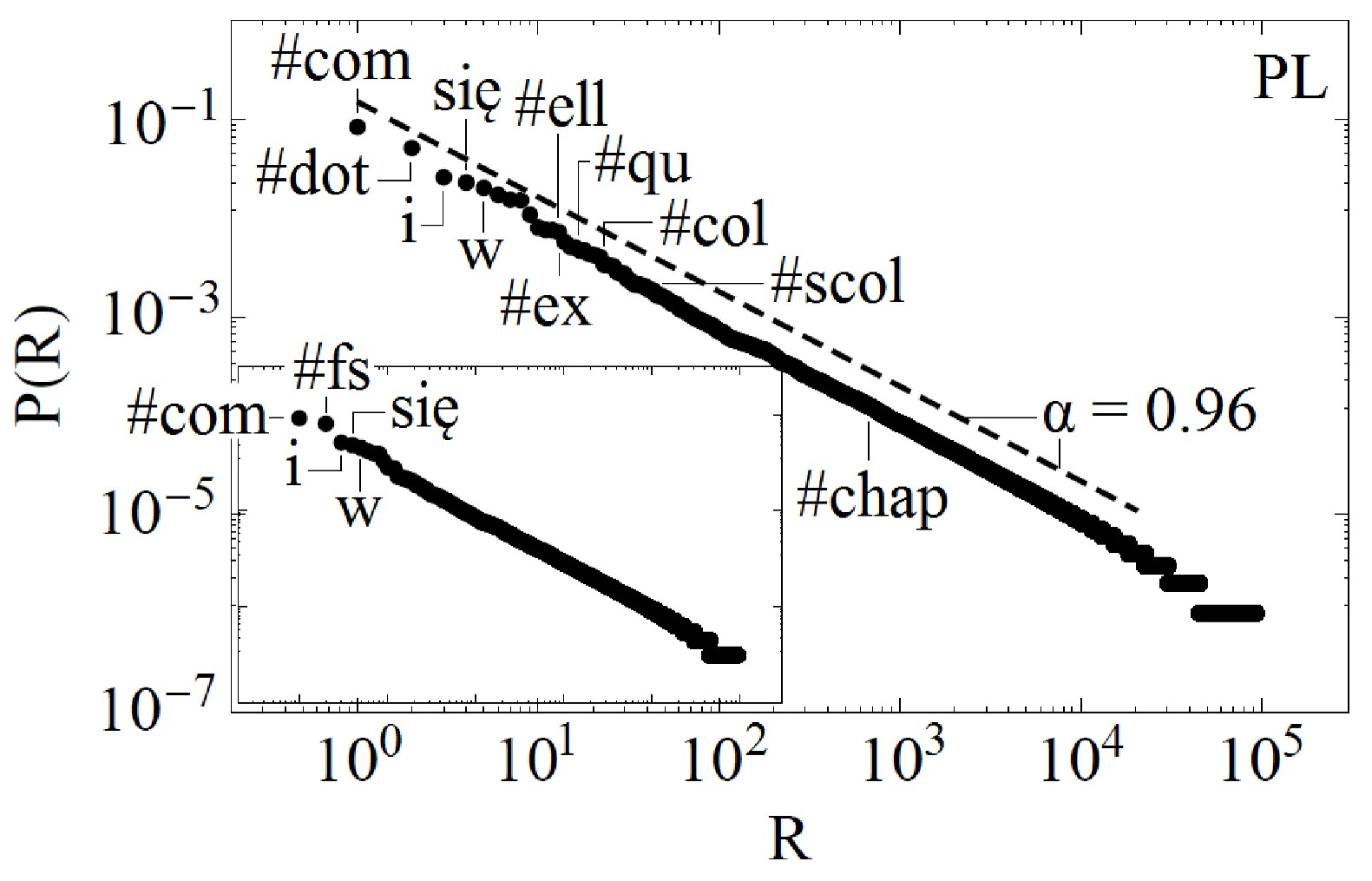}
\includegraphics[scale=0.44]{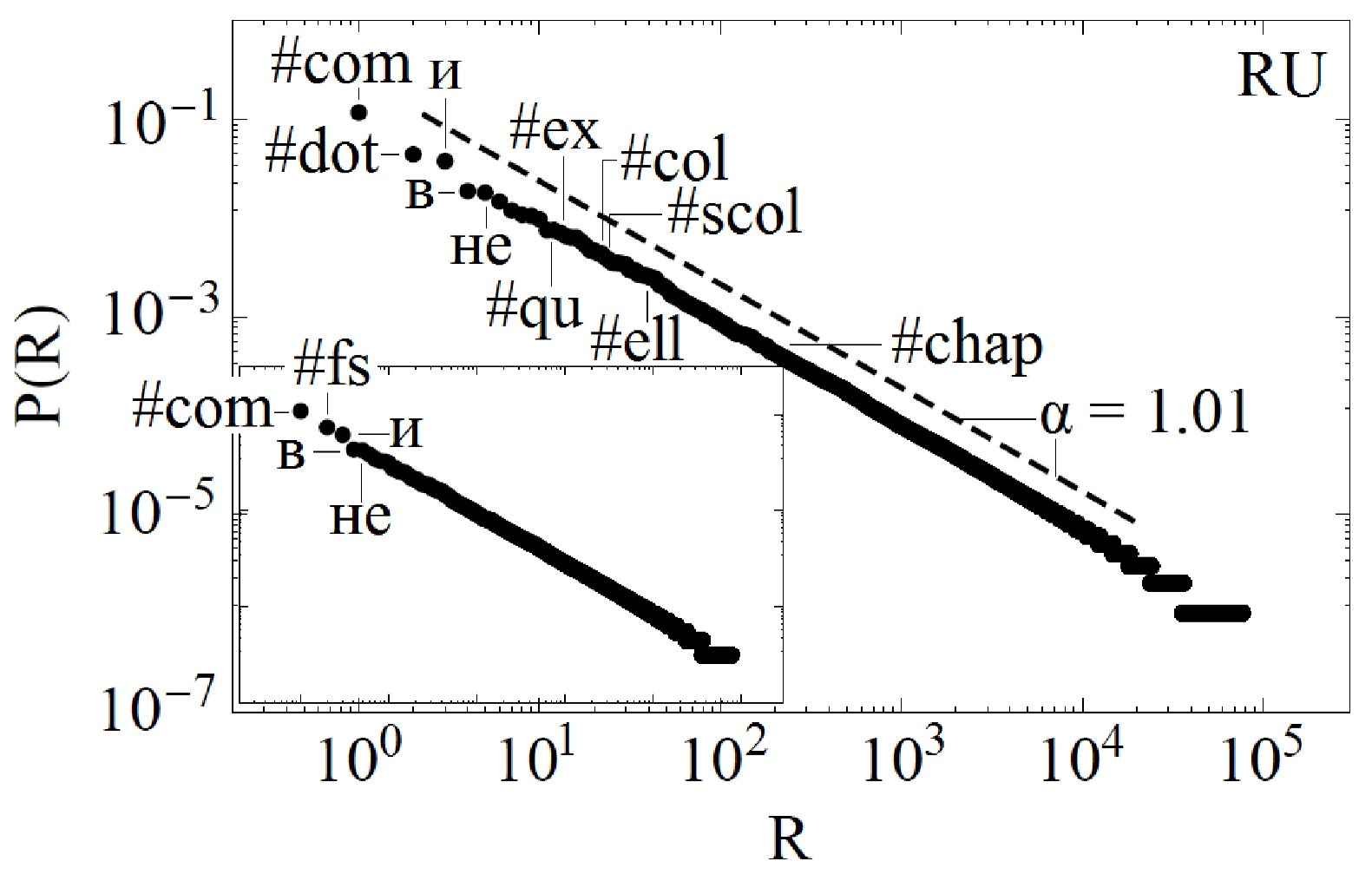}
\caption{The word and punctuation-mark occurrence probability distributions for corpora representing different European languages: English (top left), German (top right), French (middle left), Italian (middle right), Polish (bottom left), and Russian (bottom right). Each language is represented by a corpus of length $s_c \approx 10^6$ words and punctuation marks created from a set of novels. For each language, dashed lines represent the Zipfian power law fitted within the range [$10^1,10^4$] (and extended over the range $R < 10$) and a value of the related exponent $\alpha$. (Main) Different punctuation marks are counted separately: comma (\#com), dot (\#dot), question mark (\#qu), exclamation mark (\#ex), ellipsis (\#ell), semicolon (\#scol), colon (\#col), and new chapter (\#chap). (Inset) All the punctuation marks that end sentences are counted together as full stops (\#fs). In both panels the most frequent words are captioned.}
\label{fig::zipf}
\end{figure}

After calculating the frequency of words, a set of words that are present in almost every sample is selected. As it turns out, for a sufficiently large sample they are always the words having grammatical functions. Regardless of the topics covered by a sample text, these words occupy the first ranks in the Zipf distribution. Additionally, we count the occurrence numbers of different punctuation marks in each sample and include them in the corresponding Zipf distributions as if they were ordinary words. The main plots in Fig.~\ref{fig::zipf} show such distributions with distinguished punctuation marks (the special division): dot (\#dot), question mark (\#qu), exclamation mark (\#ex), ellipsis (\#ell), semicolon (\#scol), colon (\#col), comma (\#com), and new chapter (\#chap). In the insets to Fig.~\ref{fig::zipf}, all the marks that can end sentences are counted together as full stops (\#fs).

\begin{figure}[!ht]
\centering
\includegraphics[scale=0.44]{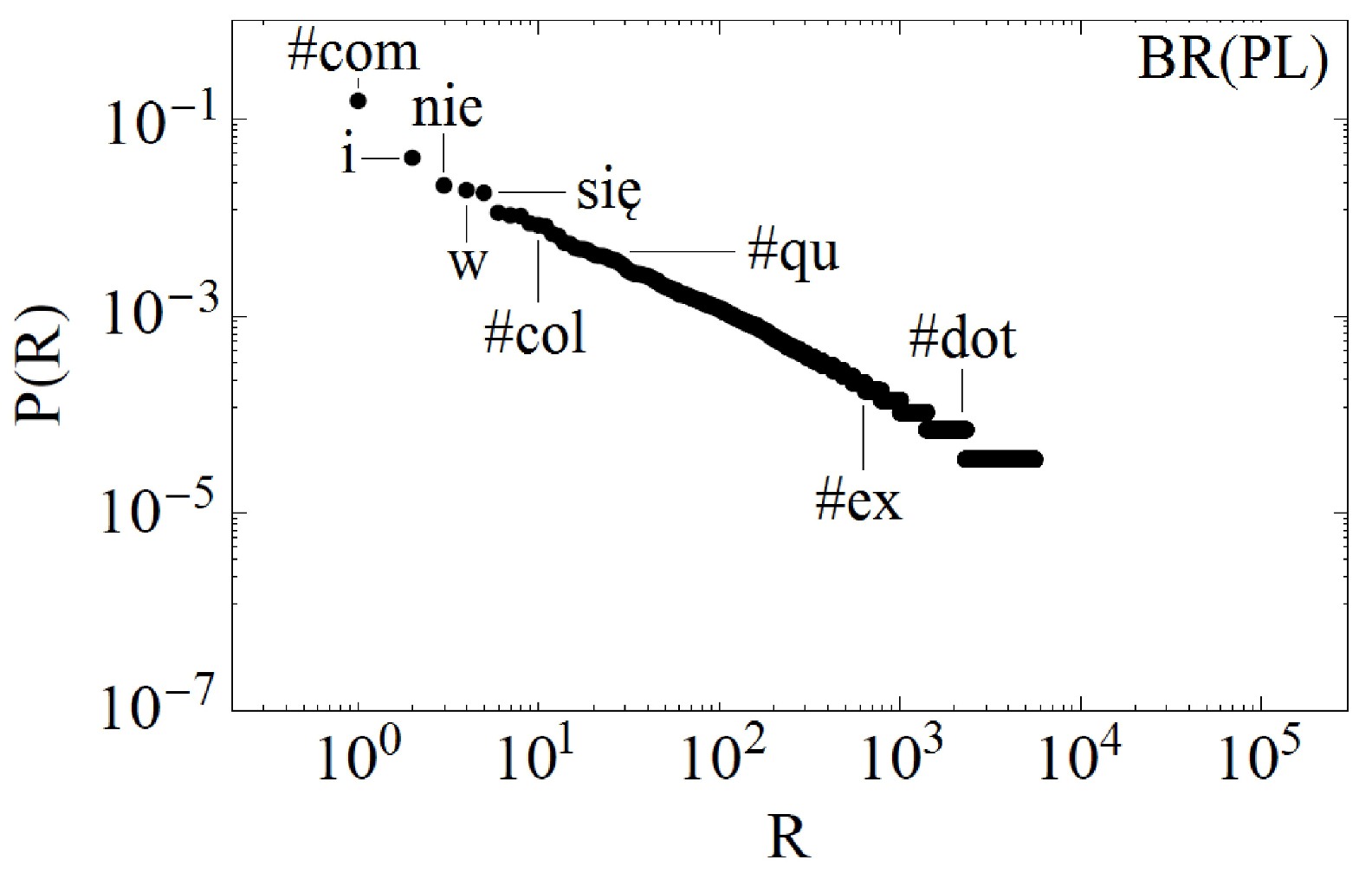}
\includegraphics[scale=0.44]{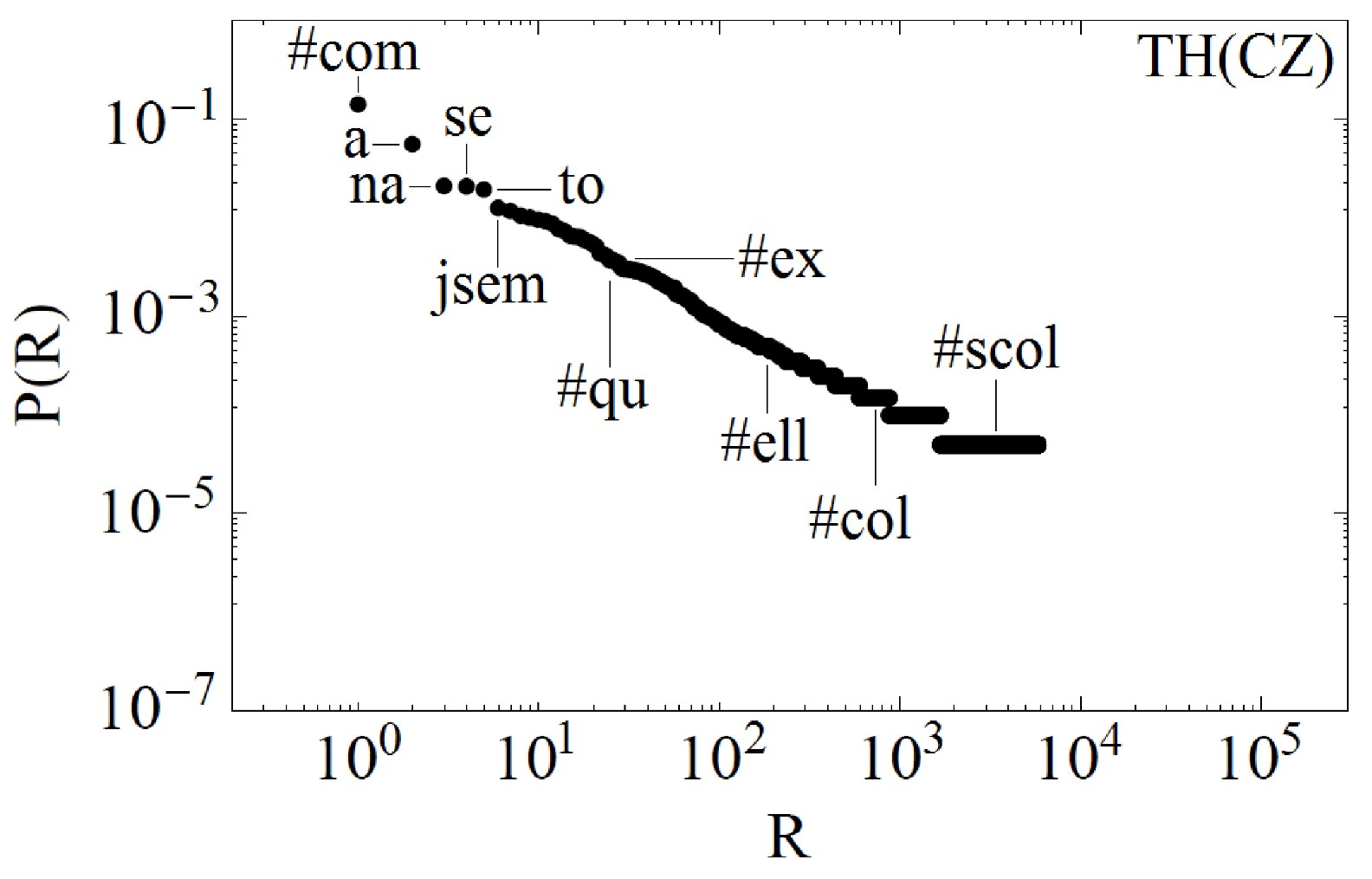}
\caption{The word and punctuation mark occurrence probability distributions for two texts with unusual punctuation statistics: {\it Bramy raju} by Jerzy Andrzejewski (Polish, left) and {\it Taneční hodiny pro starší a pokročilé} by Bohumil Hrabal (Czech, right). The abbreviations are the same as in Fig.~\ref{fig::zipf}.}
\label{fig::zipf.odd}
\end{figure}

Commas and the different types of full stops (except for ellipses) appear in the same region of the Zipf distribution where the highest-ranked words reside, i.e., the function words, like conjunctions (especially in the Slavic languages), articles (the Romance and Germanic languages), and prepositions. In all the considered languages, comma has $R=1$, while the rank of dot is typically $R=2$, except for Italian and English ($R=3$). The question and exclamation marks as well as semicolons and colons have considerably lower ranks that vary among the languages but in general can be found in the interval $10 < R < 30$ (\#qu and \#ex) and in the interval $10 < R < 50$ (\#scol and \#col). Ellispes can behave as lexical words with their ranks sometimes being lower than $R=100$. This refers even more to the new chapter marks whose frequency varies strongly from text to text and their rank can be as low as $R \approx 1000$ for particular books. For the general division, the unified full stop becomes the second most frequent object after comma in all languages except for English, where it occupies rank $R=3$ (after comma and {\it the}). The most interesting observation regarding the plots is that all the punctuation marks in both divisions are placed together with the regular words in the regime that is close to a power-law. This means that adding the punctuation marks to the Zipf analysis results in a substantial improvement of the scaling of the rank-frequency plots in that part ($R < 10$) that in a standard analysis deviates from a power-law towards the lower frequencies and that is described by the Zipf-Mandelbrot distribution. From this point of view, the punctuation marks act towards restoring of the Zipf distribution. This effect can be seen in Fig.~\ref{fig::zipf}, where the Zipfian power law fitted within the range [$10^1,10^4$] is geometrically extended over the highest ranks. For all the languages the corresponding points are closer to the power law and for French, Polish, and Russian they are placed exactly in the scaling regime. For a comparison, in Fig.~\ref{fig::zipf.odd} we present analogous Zipf plots for two texts where the punctuation differs from the standard pattern (a lack of the sentence structure of the texts). However, except for the distant location or even the absence of \#dot, the overall statistical properties of the remaining punctuation marks are normal.

\begin{figure}
\centering
\includegraphics[scale=0.55]{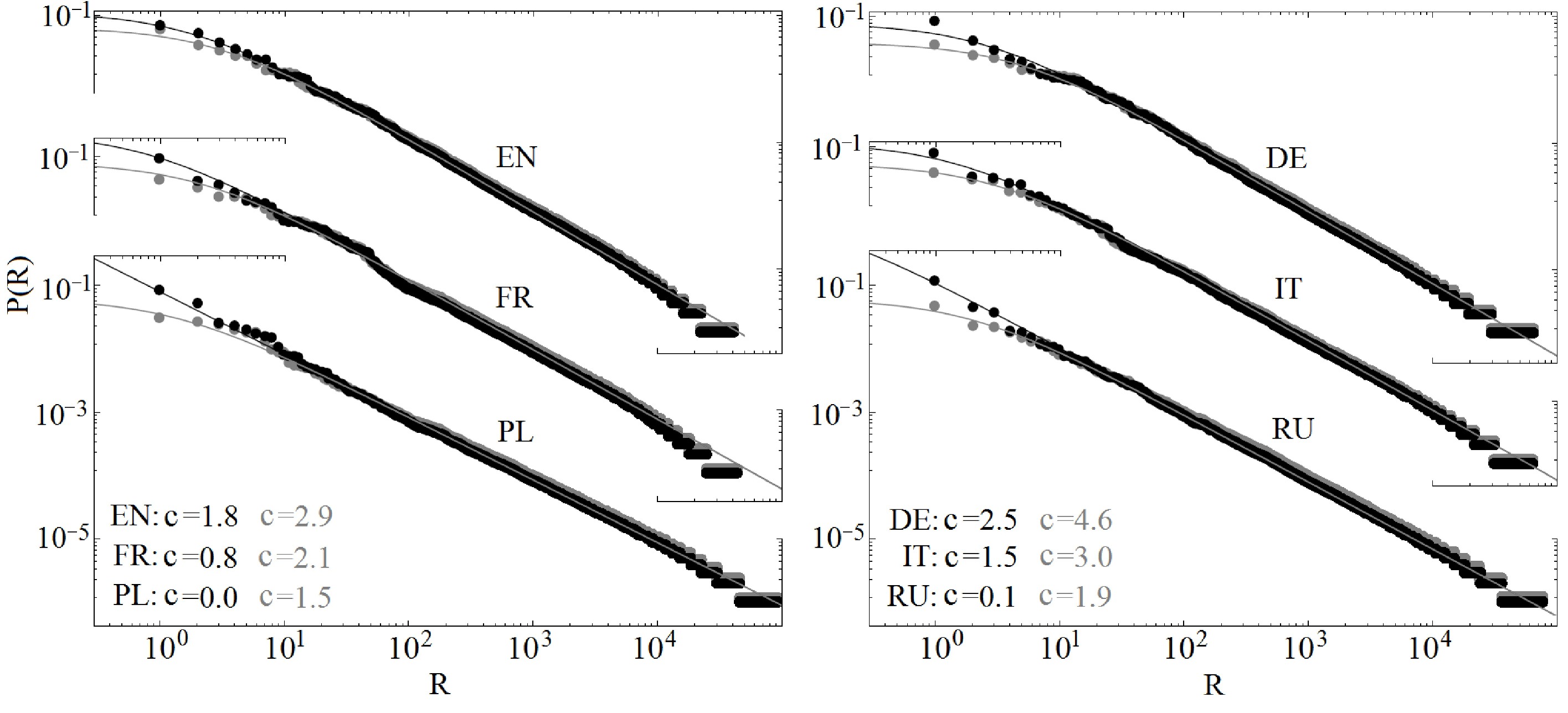}
\caption{The Zipf-Mandelbrot distribution fitted to the occurrence-probability distributions for corpora representing different European languages. The two following cases are distinguished: the words without the punctuation marks (grey) and the words with the punctuation marks (black). For each language and for both cases, the corresponding value of the best-fitted parameter $c$ is given.}
\label{fig::zipf.mandelbrot.fit}
\end{figure}

To express the above observation in a quantitative form, we fit the Zipf-Mandelbrot (Eq.~(\ref{eq::zipf.mandelbrot})) distribution to the rank-frequency plots constructed for words and for words together with the punctuation marks and estimate the corresponding values of the parameter $c$ responsible for a deflection from the pure power law $(c=0)$. Fig.~\ref{fig::zipf.mandelbrot.fit} shows such fits for all the considered languages. In each case, the inclusion of the punctuation marks results in the significantly lower values of $c$ than those in the case, in which only the words are considered, with the strength of this decrease depending on a language. It is the strongest for the Slavic languages (essentially $c=0$) and the weakest, but still sizeable, for the Germanic ones. This provides a quantitative evidence that the punctuation marks included in a Zipfian plot largely restore its scaling, indeed.

\subsection{Network properties for chosen words}

\begin{figure}[!h]
\centering
\includegraphics[scale=0.8]{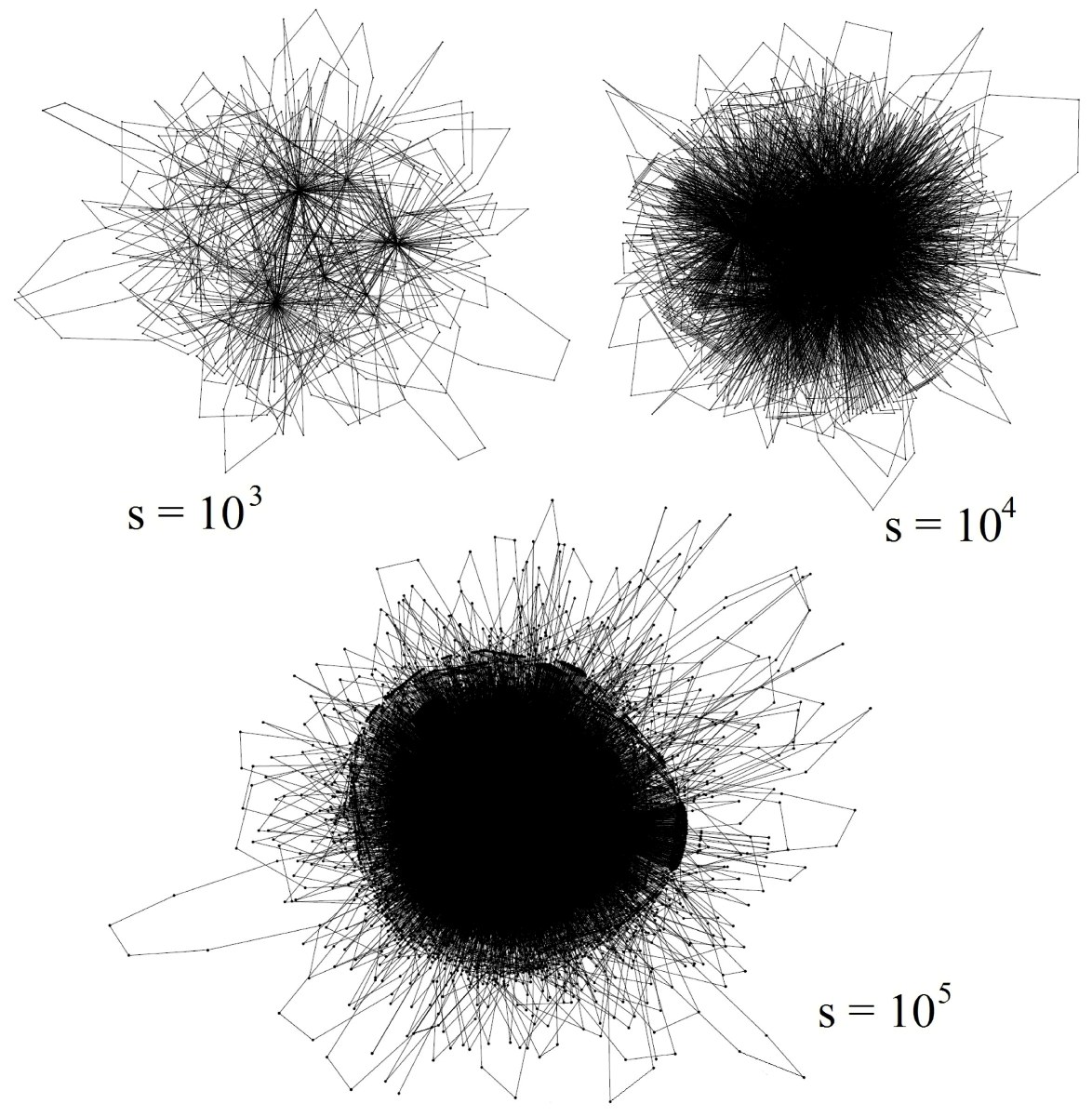}
\caption{Typical forms of a growing word-adjacency network created from text samples of length $s=10^{3},10^{4},10^{5}$ words.}
\label{fig::network}
\end{figure}

Fig.~\ref{fig::network} shows three stages of a word-adjacency network development. The network was created based on a growing sample of text of length $s$. The adopted representation allows us to check the adjacency relation between words and punctuation marks. In Tab.~\ref{table} the chosen network parameters are shown for the corpora.

\begin{table}[h!]
\centering
    \begin{tabular}{ | c | c | c | c | c | c | c | p{2cm} |}
    \hline  & English & German & French & Italian & Polish & Russian  \\ 
    \hline
    \hline
    $n$ & 40673 & 65818 & 44788 & 60985 & 89993 & 91049 \\ 
    \hline
    $e$ & 272501 & 375452 & 302243 & 398796 & 473611 & 472133 \\ 
    \hline
    \end{tabular}
\caption{Number of nodes $n$ (vocabulary size) and unique edges $e$ for word-adjacency network created based on monolingual corpora comprising $s \approx 10^{6}$ words. Since words were not lemmatized, the differences in $n$ between the languages come predominantly from inflection.}
\label{table}
\end{table}

\begin{figure}[!ht]
\centering
\includegraphics[scale=0.3]{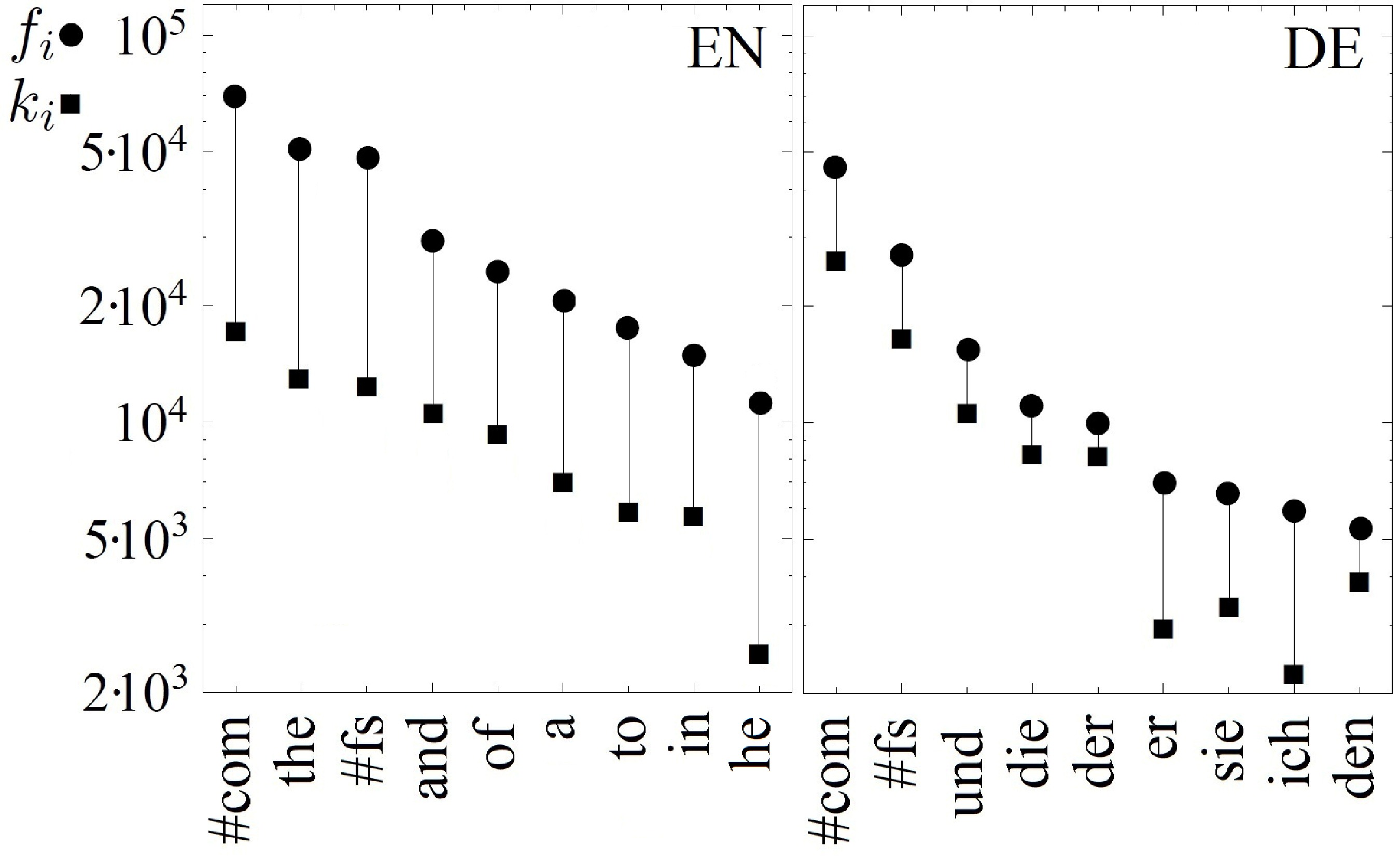}

\includegraphics[scale=0.3]{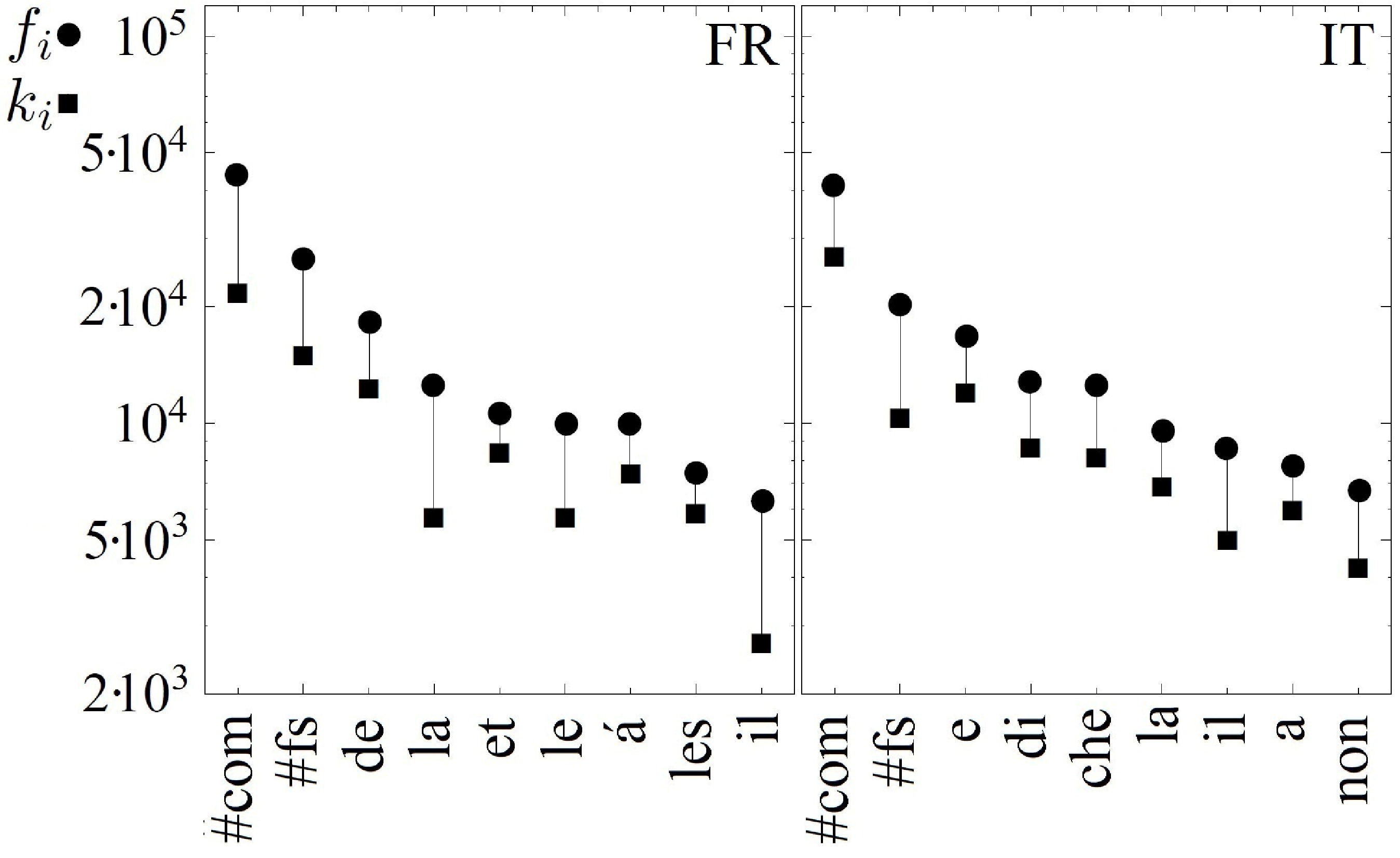}

\includegraphics[scale=0.3]{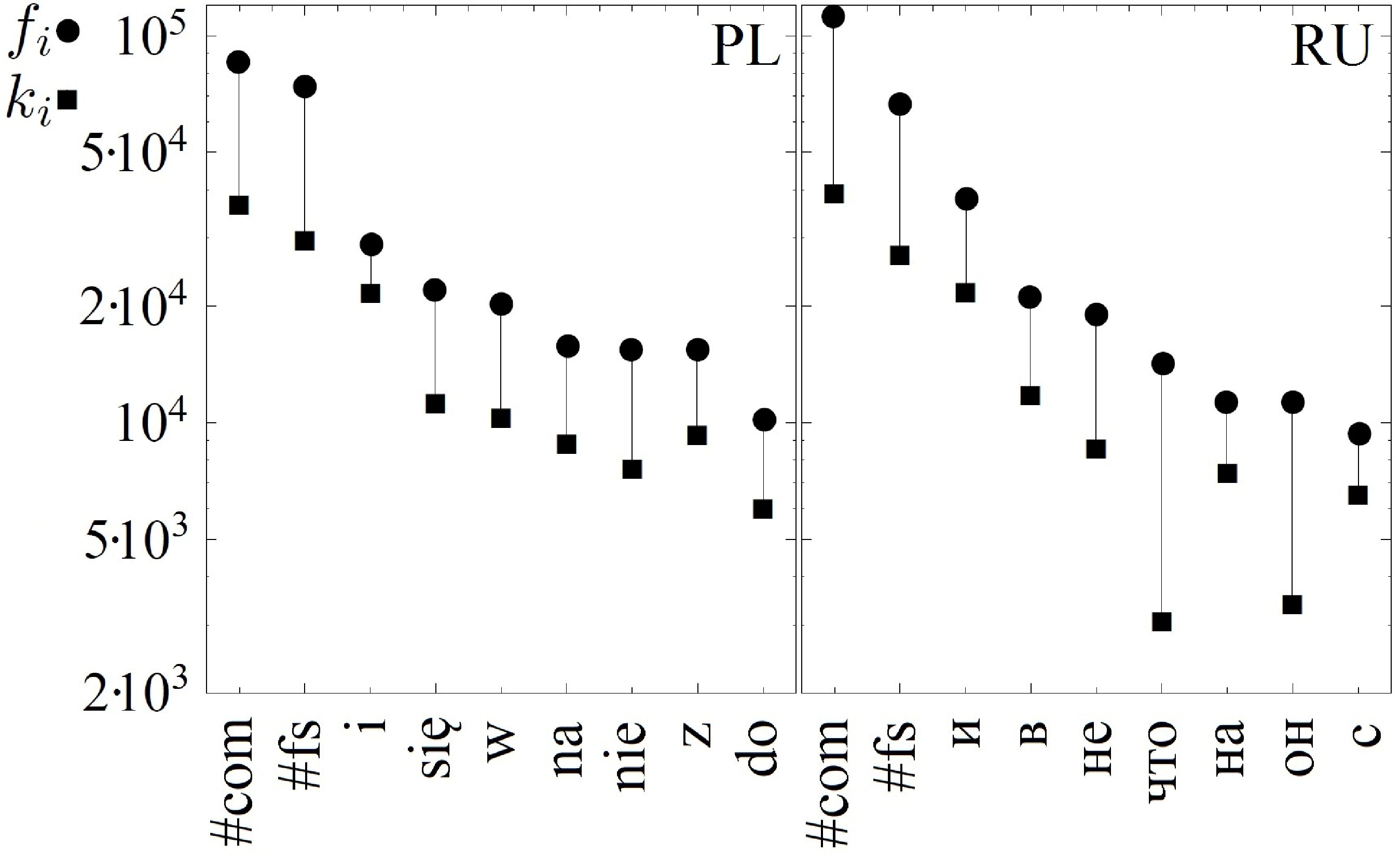}
\caption{Difference between the frequency $f_i$ of the most common words, full stops, and commas (circles) and the degree $k_i$ of the respective nodes (squares) for the Germanic (top), Romance (middle), and Slavic (bottom) languages.}
\label{fig::difference}
\end{figure}

A weighted work-adjacency network can be easily created from a text sample. The number of word co-occurrences may be understood as the weight of a connection between the respective nodes. The basic local parameter of the $i$th node is the number of edges attached to it, called a node degree $k_{i}^{w}$. It is roughly equal to doubled frequency $f_{i}$ of the corresponding word in the sample. For a binary network, a node degree $k_{i}\equiv k_{i}^{u}$ refers to the number of unique connections from the $i$th node to other nodes, where $f_{i}$ is the larger with respect to $k_{i}$, the more connections with other nodes this node has. In Fig.~\ref{fig::difference} the difference between $f_{i}$ and $k_{i}$ is shown for the most frequent items in a proper order starting from the left-hand side.

In English (Fig.~\ref{fig::difference}(top)), these differences for all the considered words are substantial and roughly similar in size on logarithmic scale. This means that there exists a simple relation: $f_i \simeq a(i) k_i$ with $1/5 < a(i) < 1/3$. The most frequent English words often form 2-grams that are repeated many times troughout the corpus, which significantly lowers the degrees of the corresponding nodes. There is also no significant difference observed between comma, full stop and the other common words. In the remaining five languages, more significant variability among different items is observed. In German, the pronouns: {\it er}, {\it sie}, and {\it ich} are represented by larger differences between both quantities ($1/6 < a(i) < 1/4$) than the punctuation marks and the other considered words ($1/3 < R < 1/2$). In French, the pronouns/articles: {\it le}, {\it la}, {\it il} show large differences up to $a(i) \approx 1/8$, the pronoun {\it les}, the preposition {\it \'a}, and the conjunction {\it et} show small differences ($a(i) \approx 3/5$), and the punctuation marks present moderate behavior (Fig.~\ref{fig::difference}(middle)). In Italian, all the considered objects except for full stop are characterized by small and steady difference between their frequency and degree. What is important, in contrast to the Germanic languages, there are comparable, rather small differences between $f_i$ and $k_i$ for the corresponding words in French and Italian.

More significant differences between $f_{i}$ and $k_{i}$ are observed for Polish and Russian (Fig.~\ref{fig::difference}(bottom)). The smallest difference is for a Polish conjunction {\it i} ($a(i) \approx 3/4$) since this word does not have any special collocation with other words. On the other hand, the punctuation marks can be collocated with specific words and this property is reflected in the largest difference between $f_i$ and $k_i$ ($a(i) \approx 1/3$), but nevertheless this difference does not exceed those observed for other words much. In Russian the variability between the items is also strong with the pronouns {\it {\foreignlanguage{russian}{что}}} and {\it {\foreignlanguage{russian}{он}}} exhibiting the largest differences ($1/6 < a(i) < 1/3$), while the conjunction {\it {\foreignlanguage{russian}{и}}} and the prepositions: {\it {\foreignlanguage{russian}{на}}}, {\it {\foreignlanguage{russian}{с}}} exhibiting the smallest ones ($a(i) \approx 3/5$). The properties of the punctuation marks in both languages are alike.

For further calculations, two other local measures are used, that is, the average shortest-path length (ASPL) for a specific node $\ell_{i}$ and the local clustering coefficient $C_{i}$. ASPL for a node $i$ refers to the average distance from a particular node to other nodes in the network and it is defined as follows:
\begin{equation}
\ell_{i}= \frac{1}{n-1} \sum_{j}^{n} d(i,j),
\end{equation}
where $d(i,j)$ denotes the shortest path (i.e., the one consisting of the minimal number of edges) between $i$ and $j$, while $n$ is the number of nodes in the network. The local clustering coefficient (LCC) for a node $i$ is:
\begin{equation}
C_{i}= \frac{2e_{i}}{k_{i}(k_{i}-1)},
\end{equation}
where $e_i$ is the number of connections between direct neighbours of the $i$th node and $k_{i}$ is its degree. This measure defines the density of links between direct neighbors of a given node and it can reveal membership of this node in a specific subset of strongly interconnected nodes~\cite{grabska2012}.
In order to calculate $\ell_i$ and $C_i$, one has to note that both quantities depend on $n$~\cite{kulig2015}. This is because, according to the Heaps law, there is a non-linear dependence between the text size $s$ and the vocabulary size $n$: $n \sim s^{\beta(s)}$ with $\beta(s)$ monotonically decreasing to zero for the infinitely long samples~\cite{gerlach2013}. In result the network becomes saturated gradually with increasing the sample size and tends to form almost a dense graph with only those edges missing that are forbidden by grammar. Therefore, typically $\ell_i$ decreases with increasing $s$, while $C_i$ increases with $s$~\cite{kulig2015}. This effect can thus be observed also in the present study if we calculate both quantities for different values of $s$ ($s \ll s_c$ in order to limit the calculation time).

Specifically, each monolingual corpus of length $s_c \approx 10^6$ was looped by connecting the last stop mark with the first word (this artificial link was removed from the networks, of course). Next, a substring of $s$ words ($10^3 \le s \le 10^5$) was randomly chosen from the corpora and transformed into a word-adjacency network (by looping the corpora, it was always possible to create a substring of words of a given length if only $s \ll s_c$). This step was repeated $m=100$ times giving a collection of $m$ networks (we allowed for the substring overlapping since, for $s \ll s_c$, obtaining two identical substrings is unlikely). The network parameters $\ell_i$ and $C_i$ were calculated for each network realization independently for the 10 most frequent items in each corpus and then their mean was also obtained: $\bar{\ell_i}= m^{-1} \sum_{m} \ell_i$ and $\bar{C_i}=m^{-1} \sum_{m} C_i$, respectively, together with its standard errors: $\sigma_{\ell_i}$ and $\sigma_{C_i}$.

\begin{figure}[t]
\centering
\includegraphics[scale=0.45]{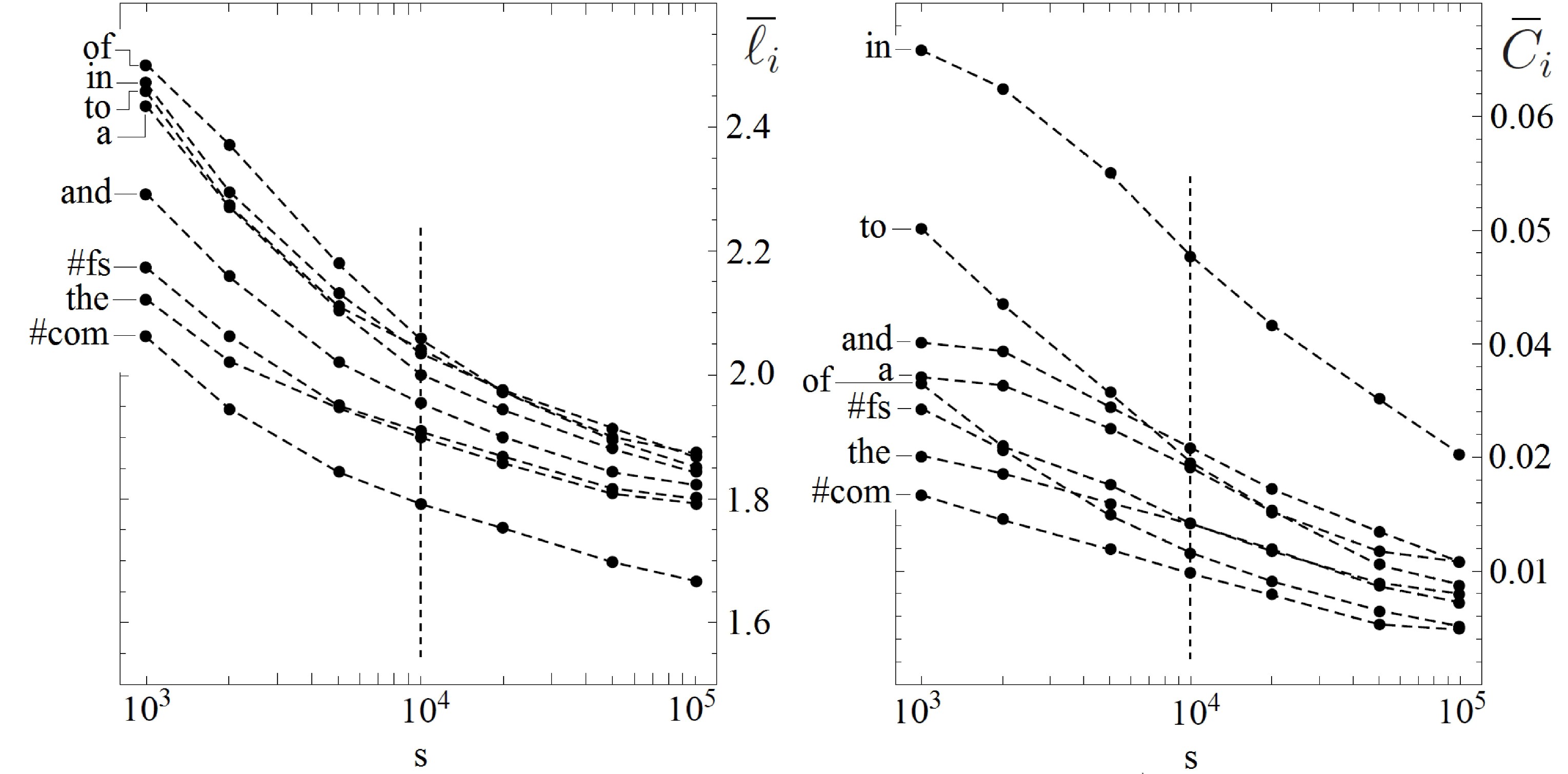}
\caption{The word-specific average shortest path length $\bar{\ell_i}$ (left) and the local clustering coefficient $\bar{C_i}$ (right) averaged over different text samples as functions of the sample size $s$ for the most frequent English words, full stops (\#fs), and commas (\#com). The vertical line indicates the value of $s=10^4$ used for creating Fig.~\ref{fig::scatter}.}
\label{fig::aspl.evo}
\end{figure}

\begin{figure}[!ht]
\centering
\includegraphics[scale=0.44]{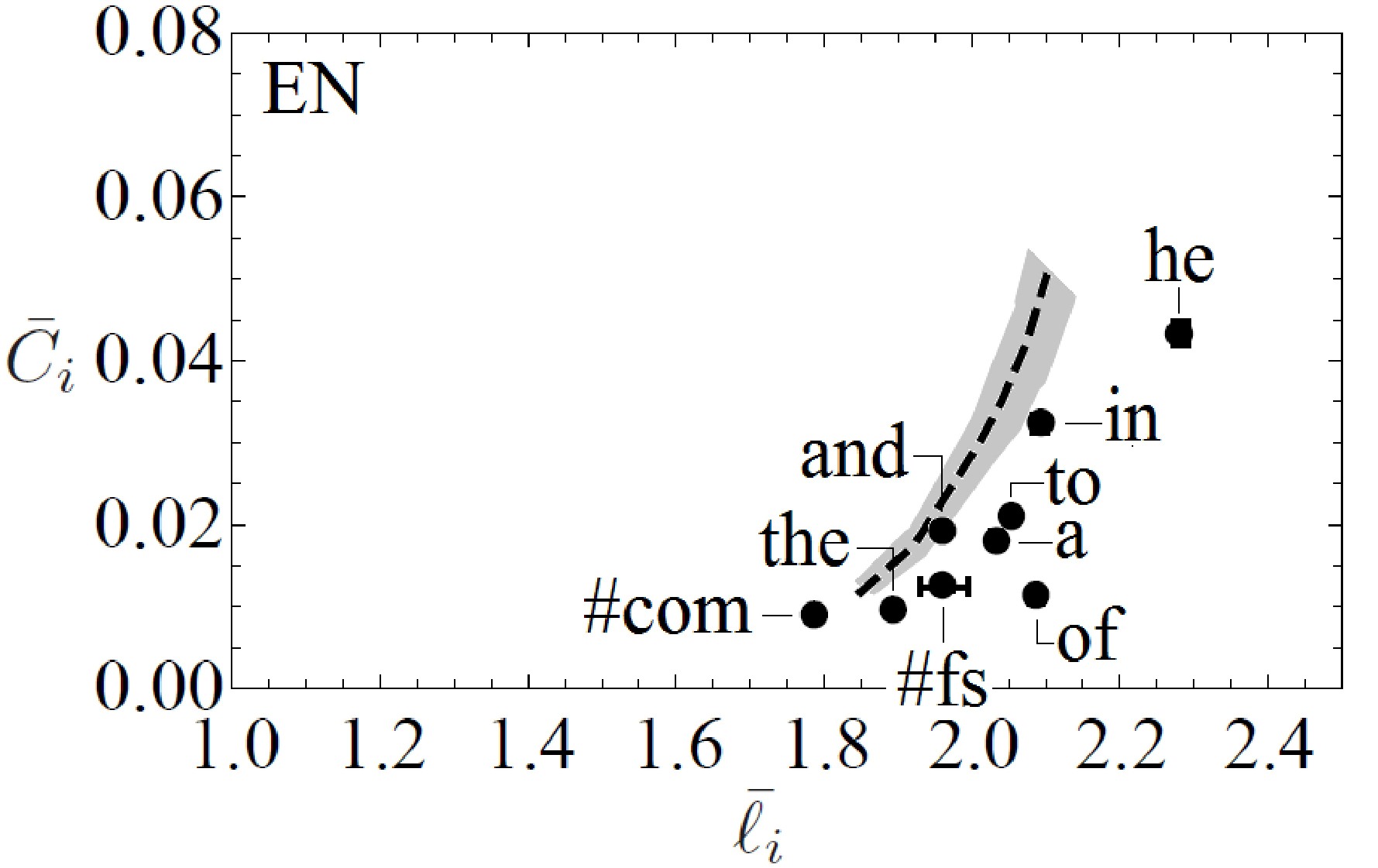}
\includegraphics[scale=0.44]{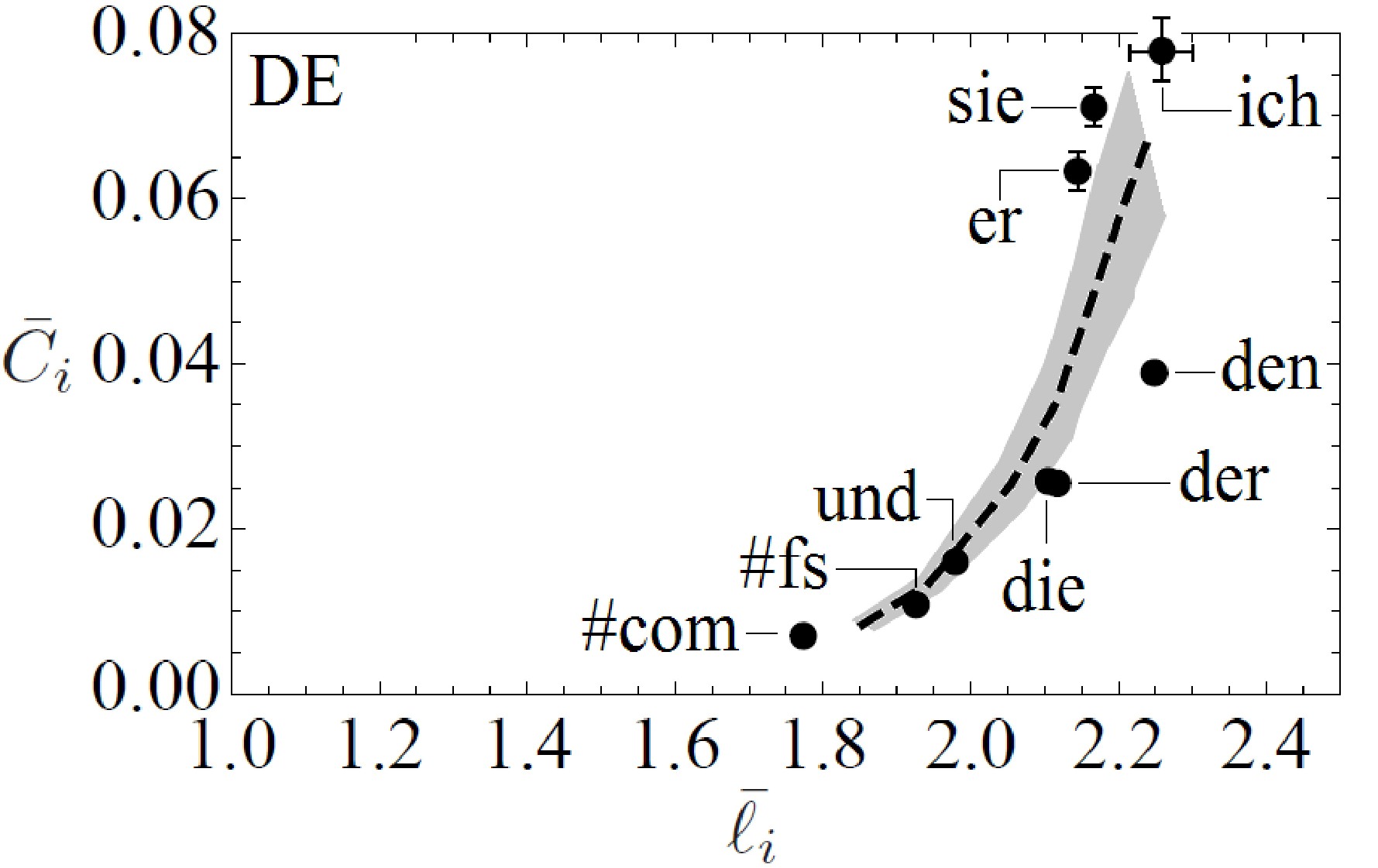}

\includegraphics[scale=0.44]{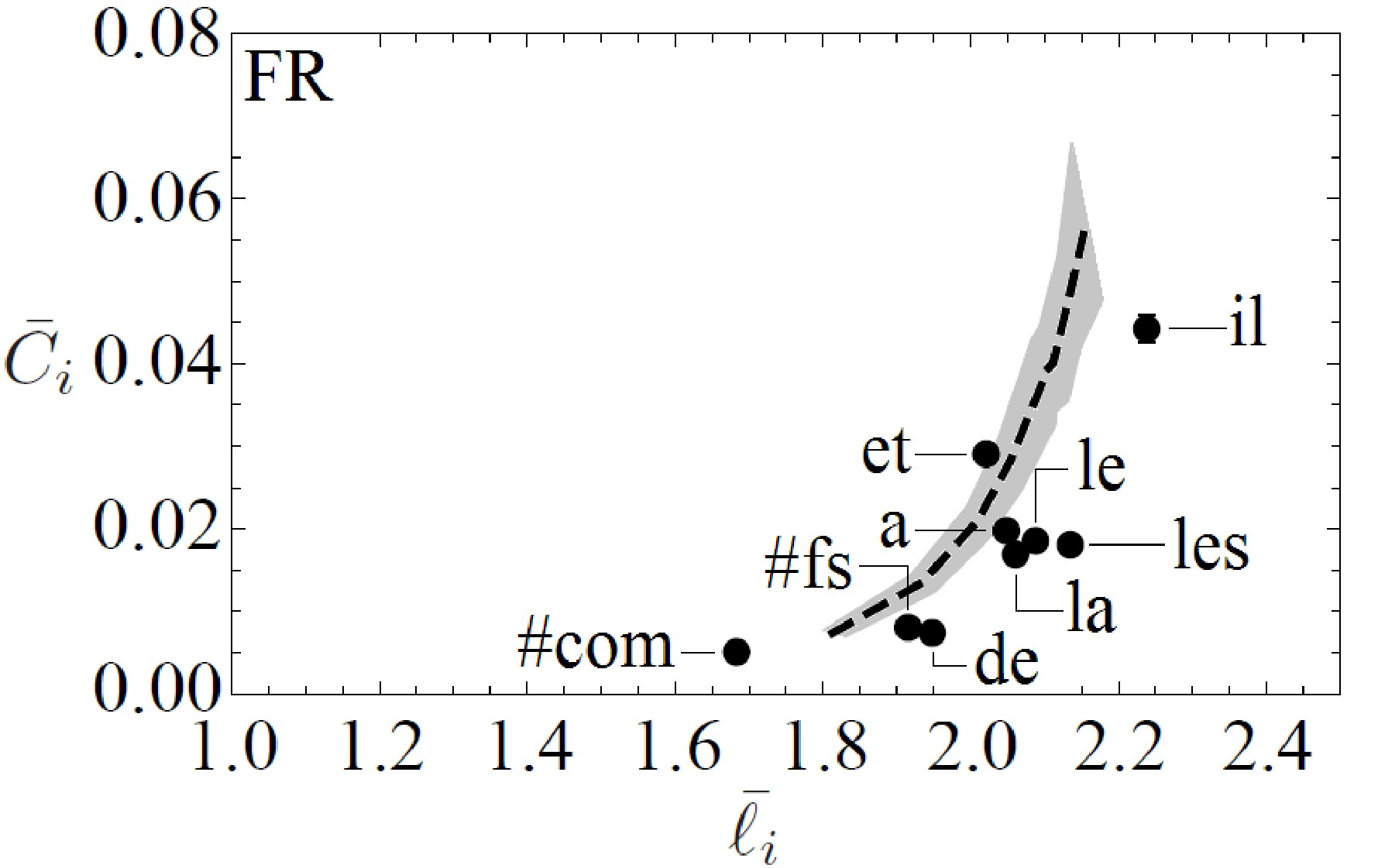}
\includegraphics[scale=0.44]{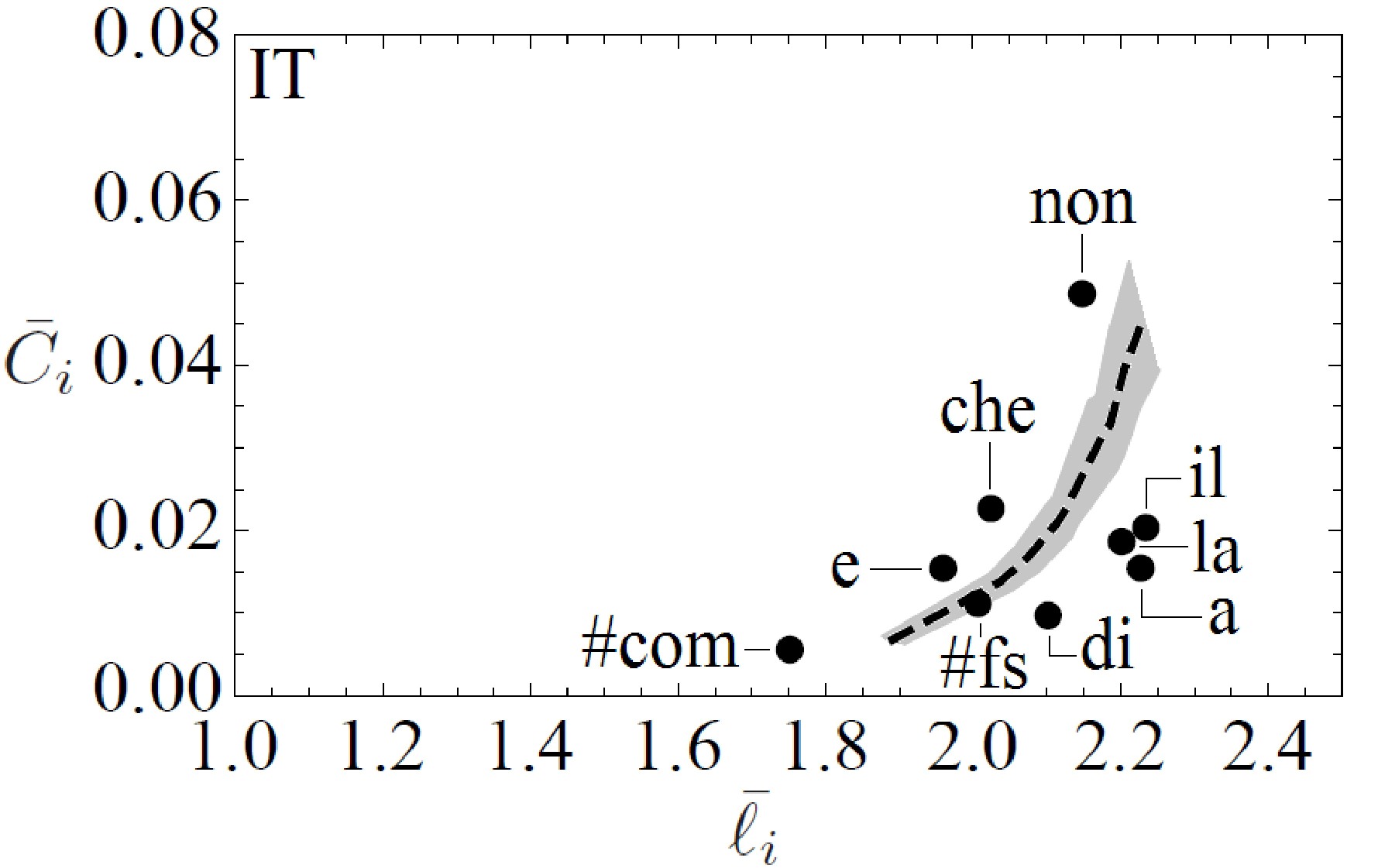}

\includegraphics[scale=0.44]{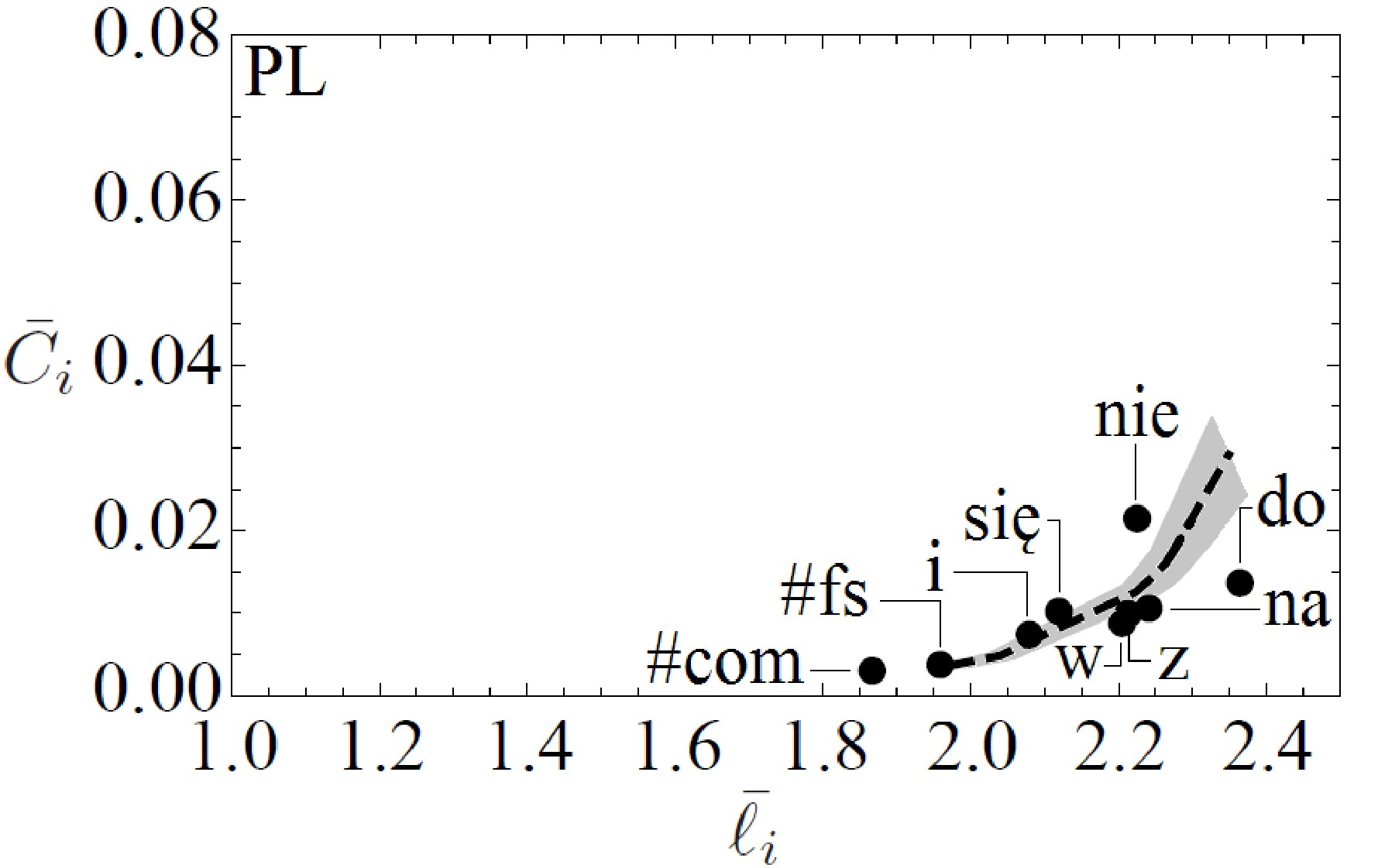}
\includegraphics[scale=0.44]{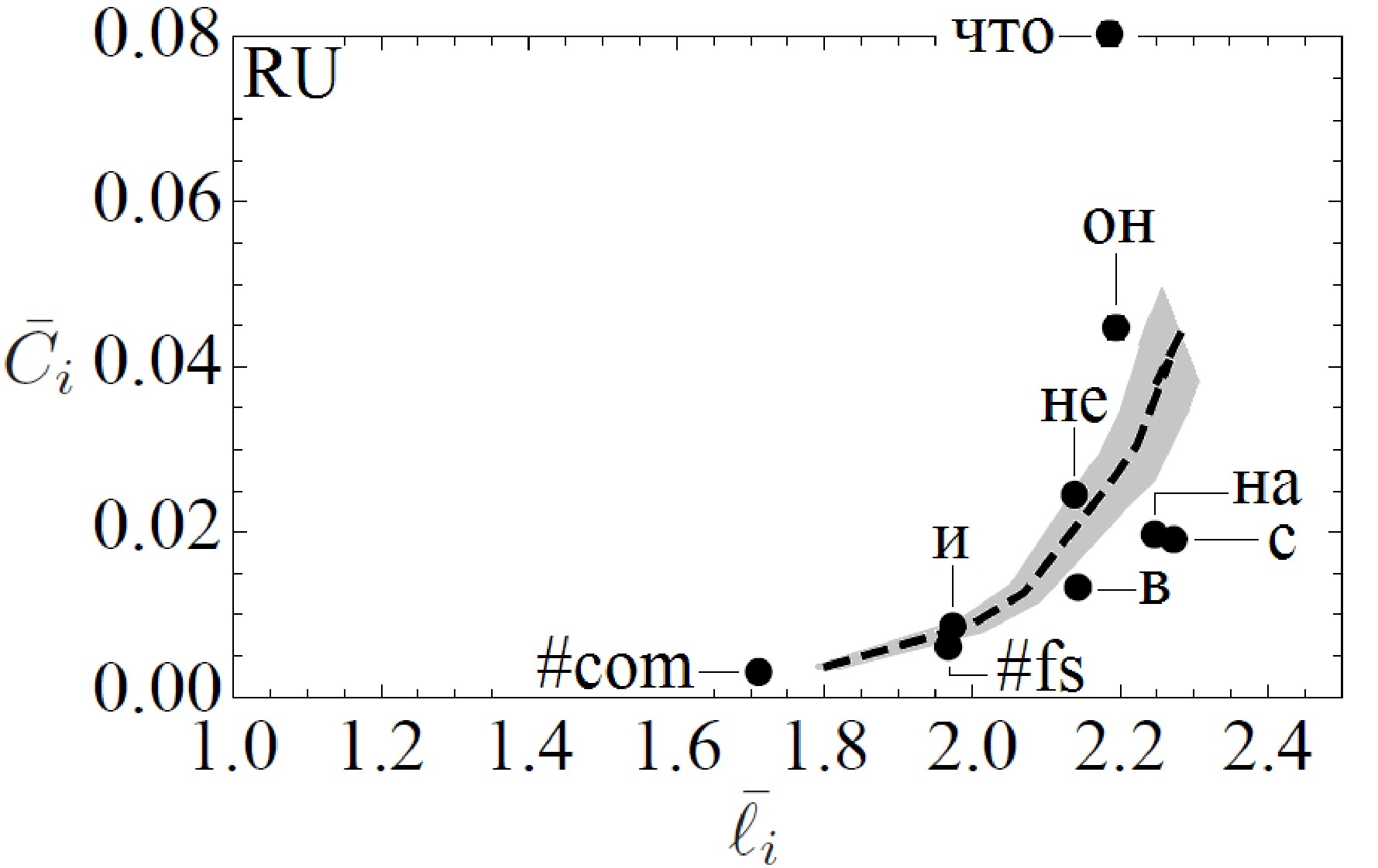}
\caption{Scatter plots of the word-specific average shortest-path length $\bar{\ell_i}$ and the local clustering coefficient $\bar{C_i}$ for the most frequent words, full stops \#fs, and commas \#com in six different European languages: English (top left), German (top right), French (middle left), Italian (middle right), Polish (bottom left), and Russian (bottom right). Error bars denote standard deviations calculated from 100 independent text samples. A null model of random word sequences (100 independent realizations) is represented by its mean (dashed line) and standard deviation (grey region).}
\label{fig::scatter}
\end{figure}

The functional dependence of $\bar{\ell}_i(s)$ and $\bar{C}_i(s)$ for the most common English words and punctuation marks is presented in Fig.\ref{fig::aspl.evo}(left) and Fig.\ref{fig::aspl.evo}(right), respectively. For the other languages considered here both plots look qualitatively similar except for that different words can be listed in each case. It is interesting to note that $\bar{\ell}_i(s)$ for \#fs and $\bar{C}_i(s)$ for both \#fs and \#com do not differ much from their counterparts representing the ordinary words, $\bar{\ell}_i(s)$ for comma is distinguished by exceptionally small values while preserving the monotonically decreasing shape of ASPL for the other objects. The results obtained for all the 6 languages are summarized in Fig.~\ref{fig::scatter} in a form of scatter plots $\bar{\ell}_i$ vs. $\bar{C}_i$ for the medium sample size of $s=10^4$. The standard errors determined for $\bar{\ell}_i$ and $\bar{C}_i$ are typically so small that they do not differ much from the symbol size in Fig.~\ref{fig::scatter}. The full stop and comma have rather low values of ASPL. Among the considered words, the most distinguished one is the German pronoun {\it ich} with a significant variability of both $\bar{\ell}_i$ and $\bar{C}_i$ among the individual sample networks. Although not explicitly shown here, the same observation refers to this word's counterparts in other languages (like {\it I}, {\it je}, {\it ja}, {\it {\foreignlanguage{russian}{я}}}), whose variability is related to particular choices of the considered texts with different narration types.

Owing to the ASPL definition, in each case the value of $\bar{\ell}_{i}$ is negatively correlated with the node degree $k_i$. LCC is also strongly anticorrelated with $k_i$ and its empirical dependence on the node degree is roughly power-law, which agrees with the theoretical considerations for the hierarchical networks~\cite{dorogovtsev2002}. This double dependence on $k_i$ means that $\bar{C}_i$ may also be considered a function of $\bar{\ell}_i$. We expect thus that substantial contribution to the variability of $\bar{C}_i$ and $\bar{\ell}_i$ in Fig.~\ref{fig::scatter} comes from this relation. In order to show this, we calculated both quantities for the random null model, in which no correlations between the words are allowed and their occurrences are governed only by their relative frequency given by the Zipf distribution. The corpora in each language was randomly shuffled, so the constituent texts lost their significance as they became just meaningless word sequences. Then we constructed the corresponding word-adjacency networks and calculated both ASPLs and LCCs for the nodes representing the same words as in Fig.~\ref{fig::scatter}. We repeated this procedure 100 times independently. Indeed, for each language we obtained an approximately power-law relation: $\bar{C}_i(\bar{\ell}_i) \sim \bar{\ell}_i^{\gamma}$ with $\gamma > 0$ (denoted by a dashed line in Fig.~\ref{fig::scatter}).

\begin{table}
\centering
    \begin{tabular}{ | c | c | c | c | c | c | c | p{2cm} |}
    \hline $R$ & English & German & French & Italian & Polish & Russian  \\ 
    \hline
    \hline
    {\it time} & 75 & 118 & 145 & 96 & 228 & 93 \\ 
    \hline
    {\it face} & 130 & 185 & 247 & 527 & 168 & 124 \\ 
    \hline
    {\it home} & 264 & 242 & 157 & 145 & 429 & 340 \\ 
    \hline
    \end{tabular}
\caption{Ranks $R$ of the lexical words used in Fig.~\ref{fig::scatter.random}.}
\label{tab::ranks}
\end{table}

\begin{figure}[!ht]
\centering
\includegraphics[scale=0.5]{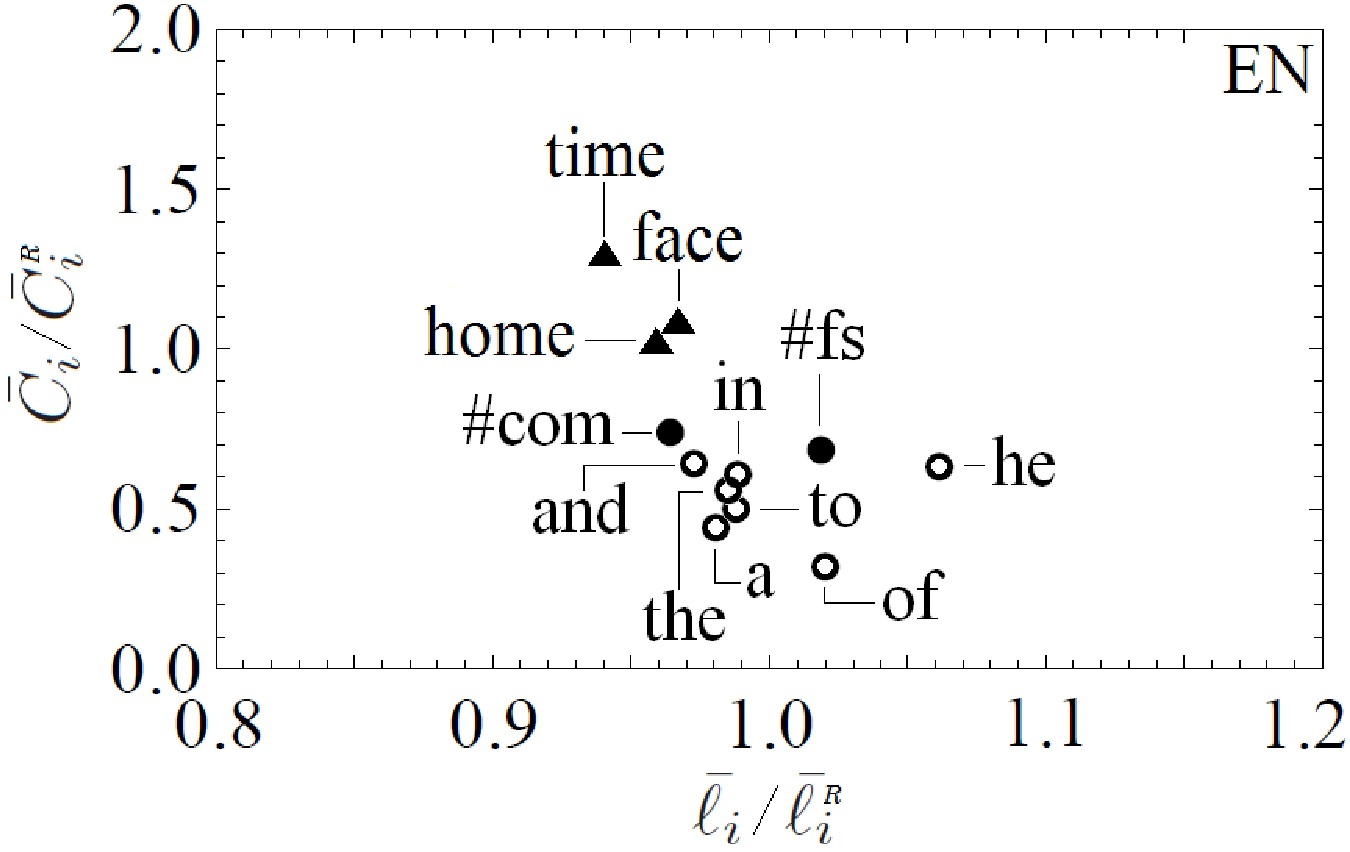}
\includegraphics[scale=0.5]{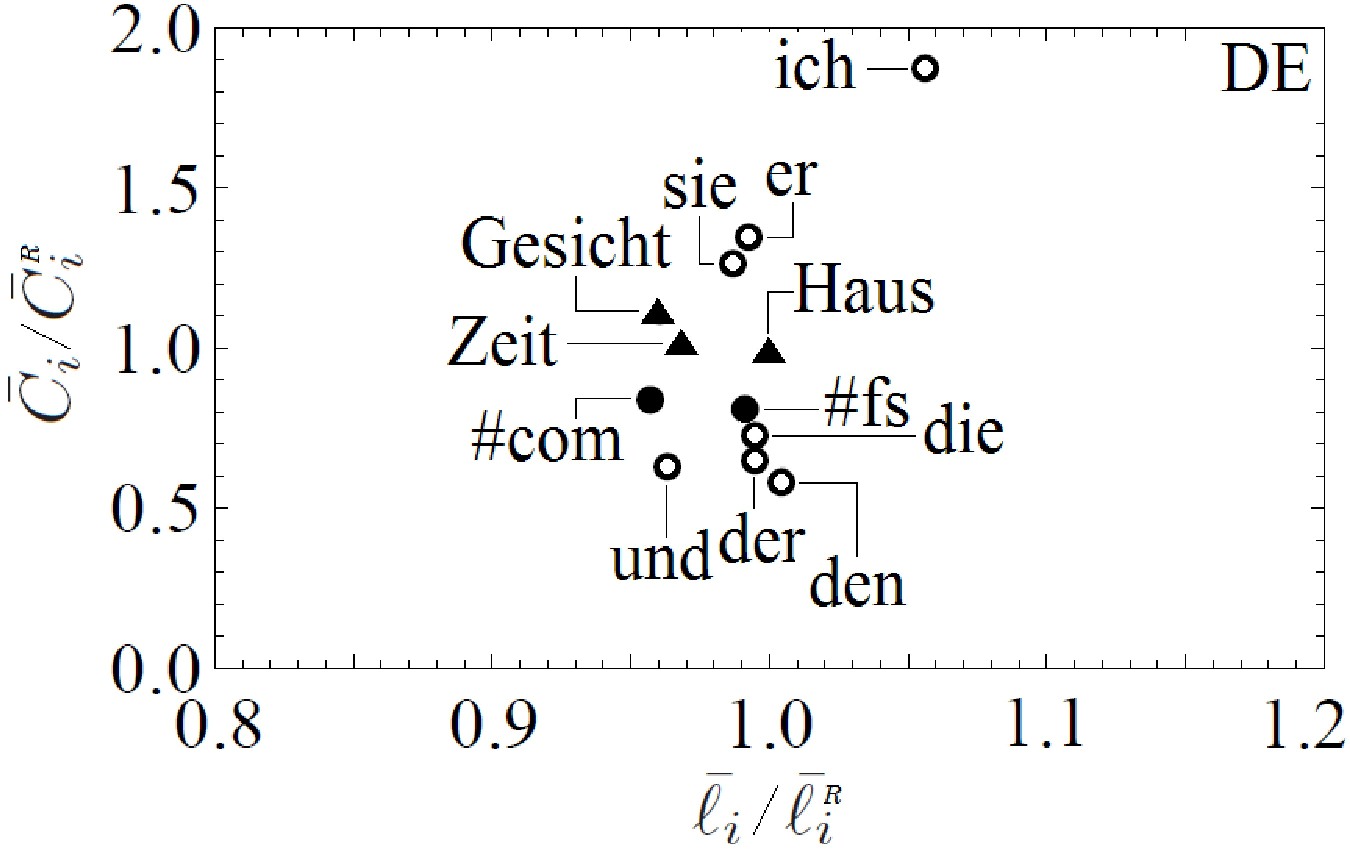}

\includegraphics[scale=0.50]{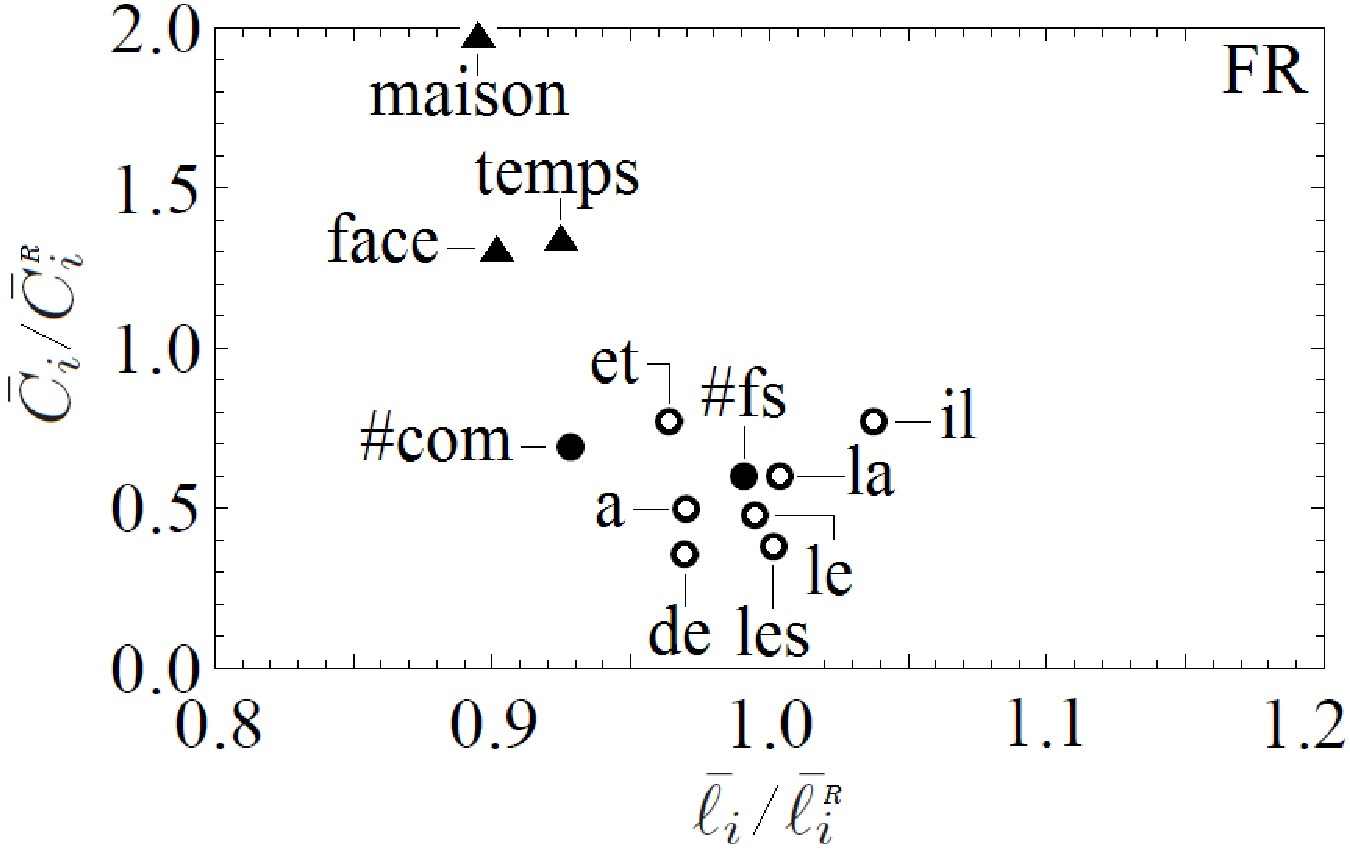}
\includegraphics[scale=0.5]{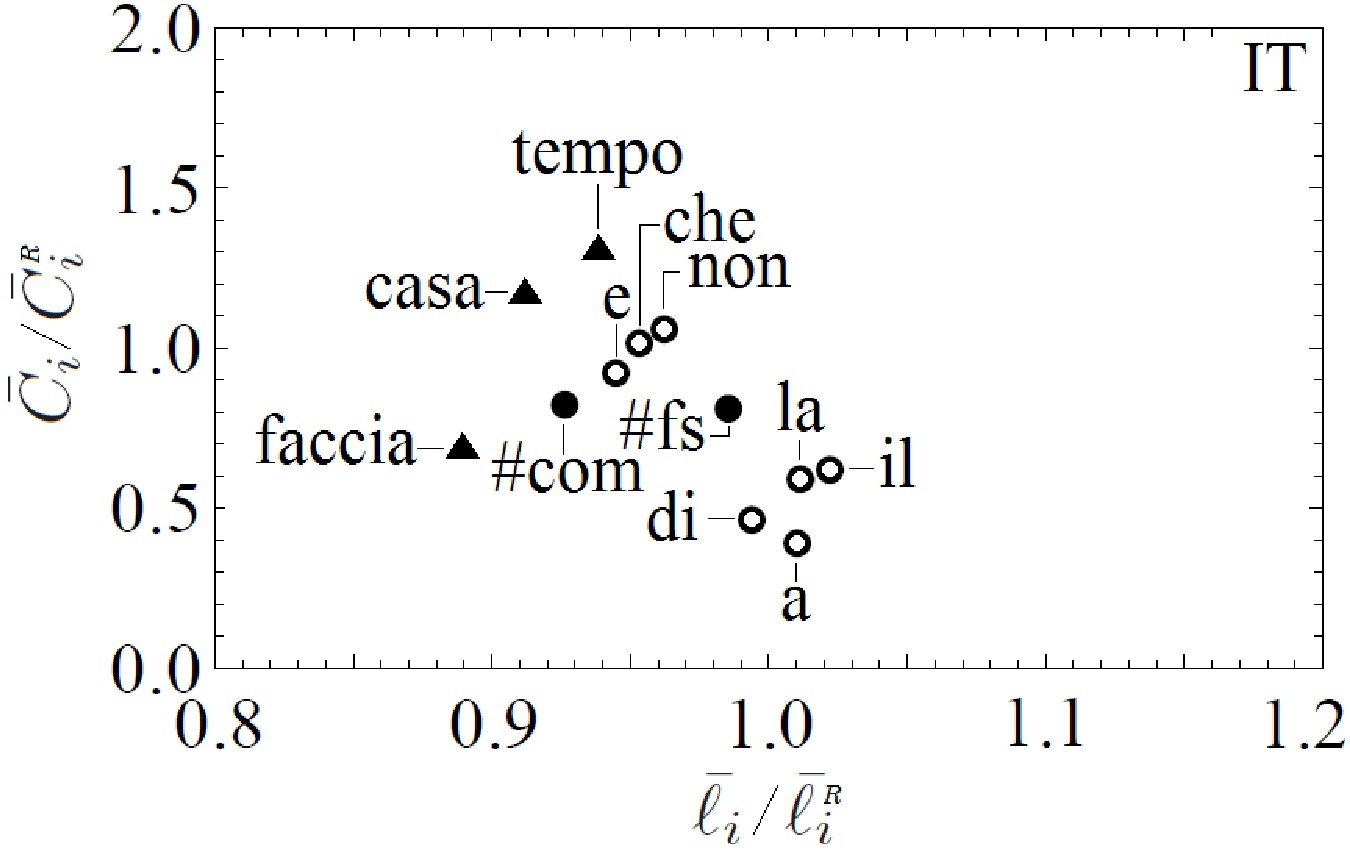}

\includegraphics[scale=0.5]{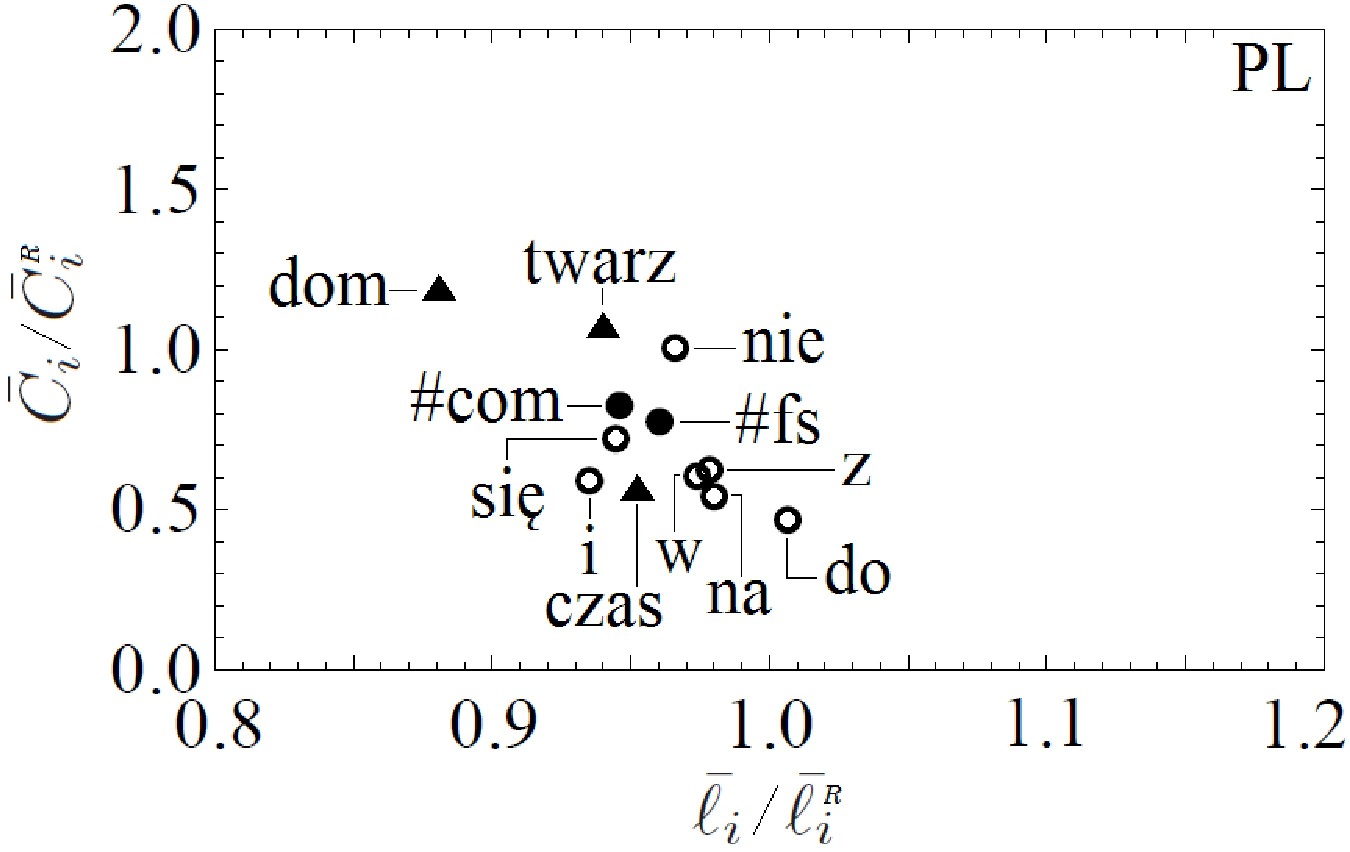}
\includegraphics[scale=0.5]{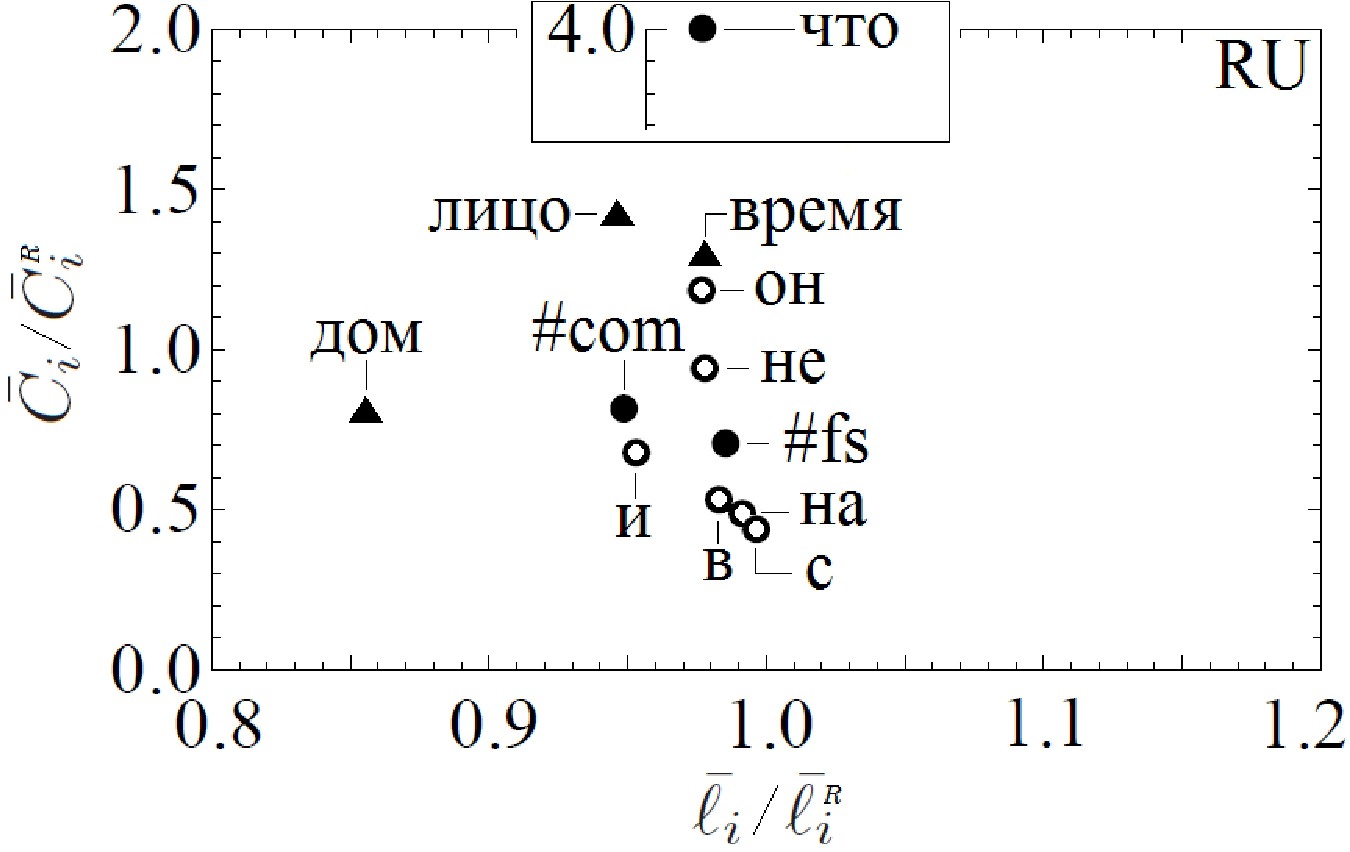}
\caption{Scatter plots of the word-specific average shortest-path length $\bar{\ell_i}$ and the local clustering coefficient $\bar{C_i}$ divided by their random-null-model counterparts $\bar{\ell}_i^{\rm R}$ and $\bar{C}_i^{\rm R}$, respectively. The same corpora as in Fig.~\ref{fig::scatter} are used. In addition to the words (open circles) and the punctuation marks (full circles) used in Fig.~\ref{fig::scatter}, the results for three sample lexical words, semantically the same for each language, are also presented (full triangles).}
\label{fig::scatter.random}
\end{figure}

The points that denote the ($\bar{\ell}_i$,$\bar{C}_i$) coordinates for the particular items in Fig.~\ref{fig::scatter} are distributed along this functional dependence. This means that the item positions on the scatter plots are strongly influenced by these items' frequencies, while the actual grammar- and context-related contributions to $\bar{C}_i$ and $\bar{\ell}_i$ are less evident. Therefore, we decided to remove the frequency-based contributions by dividing the empirical values by their average random-model counterparts: $\bar{C}_i^{\rm R}$ and $\bar{\ell}_i^{\rm R}$. The resulting positions of the items are shown in Fig.~\ref{fig::scatter.random}. Now it is more evident than in Fig.~\ref{fig::scatter} that both the high-frequency words and the punctuation marks occupy similar positions and no quantitative difference can be identified that is able to distinguish between both groups. In this figure, we also show these quantities calculated for three sample words chosen randomly from more distant parts of the Zipf plot: {\it time} ($R=75$ in the English corpus), {\it face} ($R=130$), and {\it home} ($R=264$), as well as their semantical counterparts in the other languages (occupying different ranks, see Tab.~\ref{tab::ranks}). Obviously, each of these words may also have other, non-equivalent meanings in distinct languages and, while some languages use inflection, the other ones do not, which inevitably contribute to the rank differences. In contrast to the most frequent words discussed before, these words are significantly less frequent, which can itself lead to some differences in the statistical properties as compared to the top-ranked words. Therefore, they are not shown in Fig.~\ref{fig::scatter}, because their local clustering coefficient significantly exceeds the vertical axis upper limit. In Fig.~\ref{fig::scatter.random}, the sample lexical words are located in different places for different languages, but typically their average position is more or less shifted towards the upper left corner of the plots. This effect is the most pronounced for French, then for English, Russian, Italian, and Polish, while it is absent for German. This visible shift may originate from either the statistical fluctuations among the words, the statistical fluctuations among the texts selected for the corpora, or be a geniune effect for the less frequent words, the parts of speech, and/or a general property of the lexical words in specific languages. However, since our sample of the medium-ranked words is small, at present we prefer not to infer any decisive conclusions from this result as we plan to carry out a related, comprehensive study in near future. Nevertheless, we stress here that such displacements exhibited in Fig.~\ref{fig::scatter.random} by the words of medium frequency by no means contradict our main statement that the punctuation marks show similar statistical properties as the most frequent words.

\begin{figure}[!h]
\centering
\includegraphics[scale=0.88]{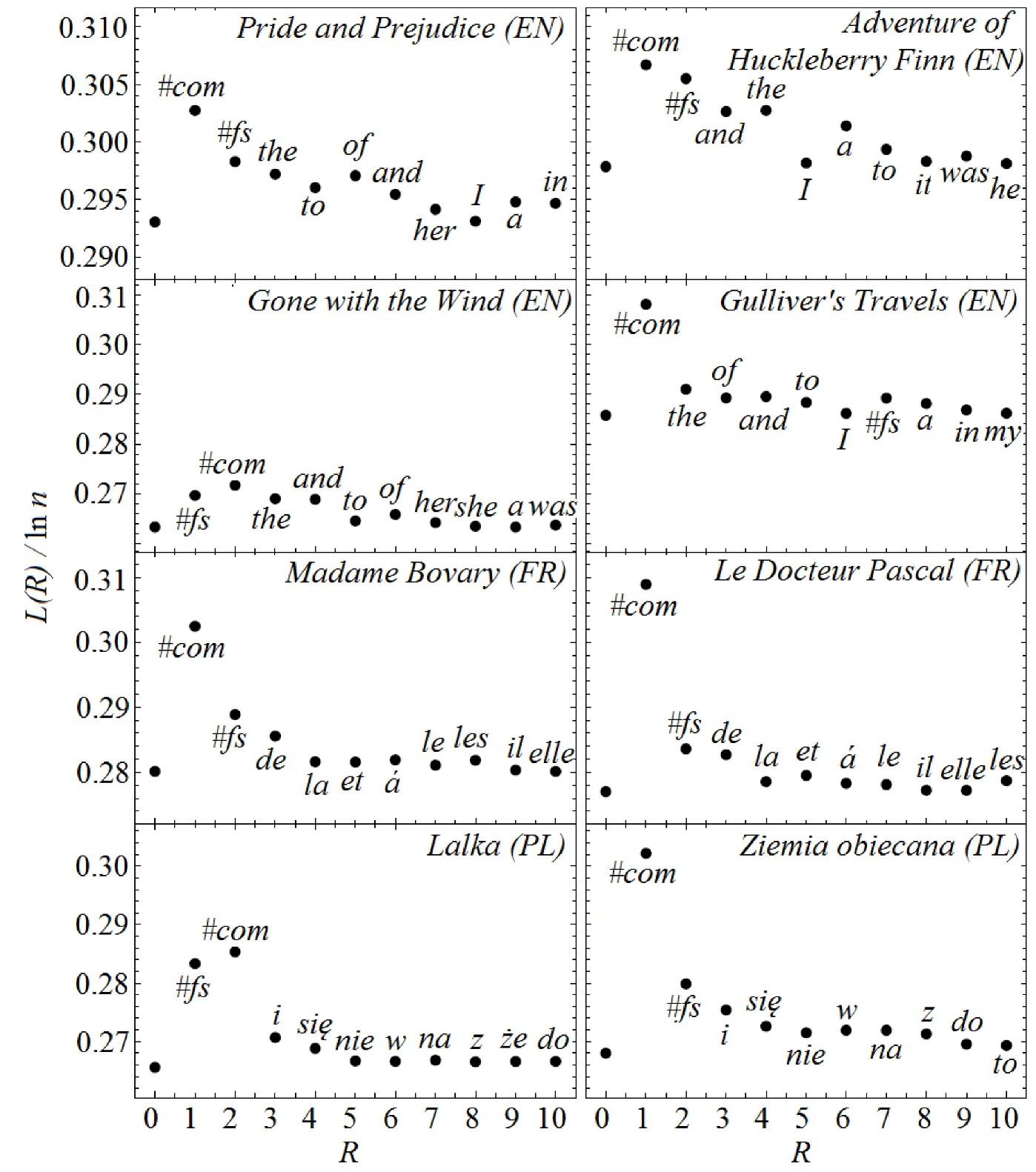}
\caption{The average shortest path length $L(R)$ with respect to $\ln n$ for the networks representing different text samples (novels). For each novel, the rank $R=0$ denotes the complete network with all the $n$ nodes, while the lower ranks $1 \le R \le 10$ denote the incomplete networks with $n-1$ nodes obtained by removing a node corresponding to a word ranked $R$ in the Zipf distribution for this novel.}
\label{fig::aspl.removed}
\end{figure}

\begin{figure}[!h]
\centering
\includegraphics[scale=0.88]{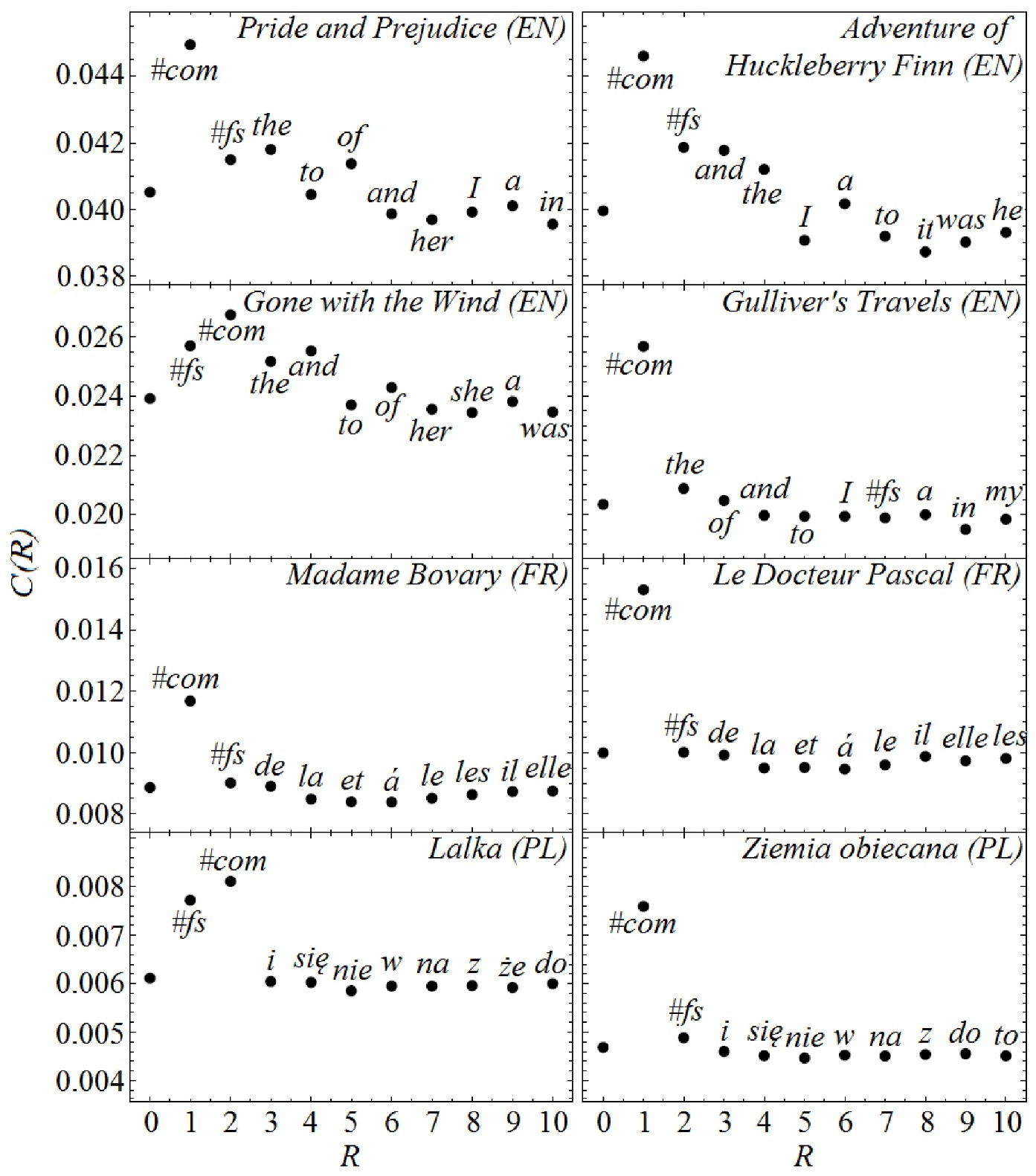}
\caption{The global clustering coefficient $C(R)$ for the networks representing different novels. For each novel, the rank $R=0$ denotes the complete network with all $n$ nodes, while the lower ranks $1 \le R \le 10$ denote the incomplete networks with $n-1$ nodes obtained by removing a node corresponding to the word ranked $R$ in the Zipf distribution for this novel.}
\label{fig::gcc.removed}
\end{figure}

\begin{figure}[!h]
\centering
\includegraphics[scale=0.88]{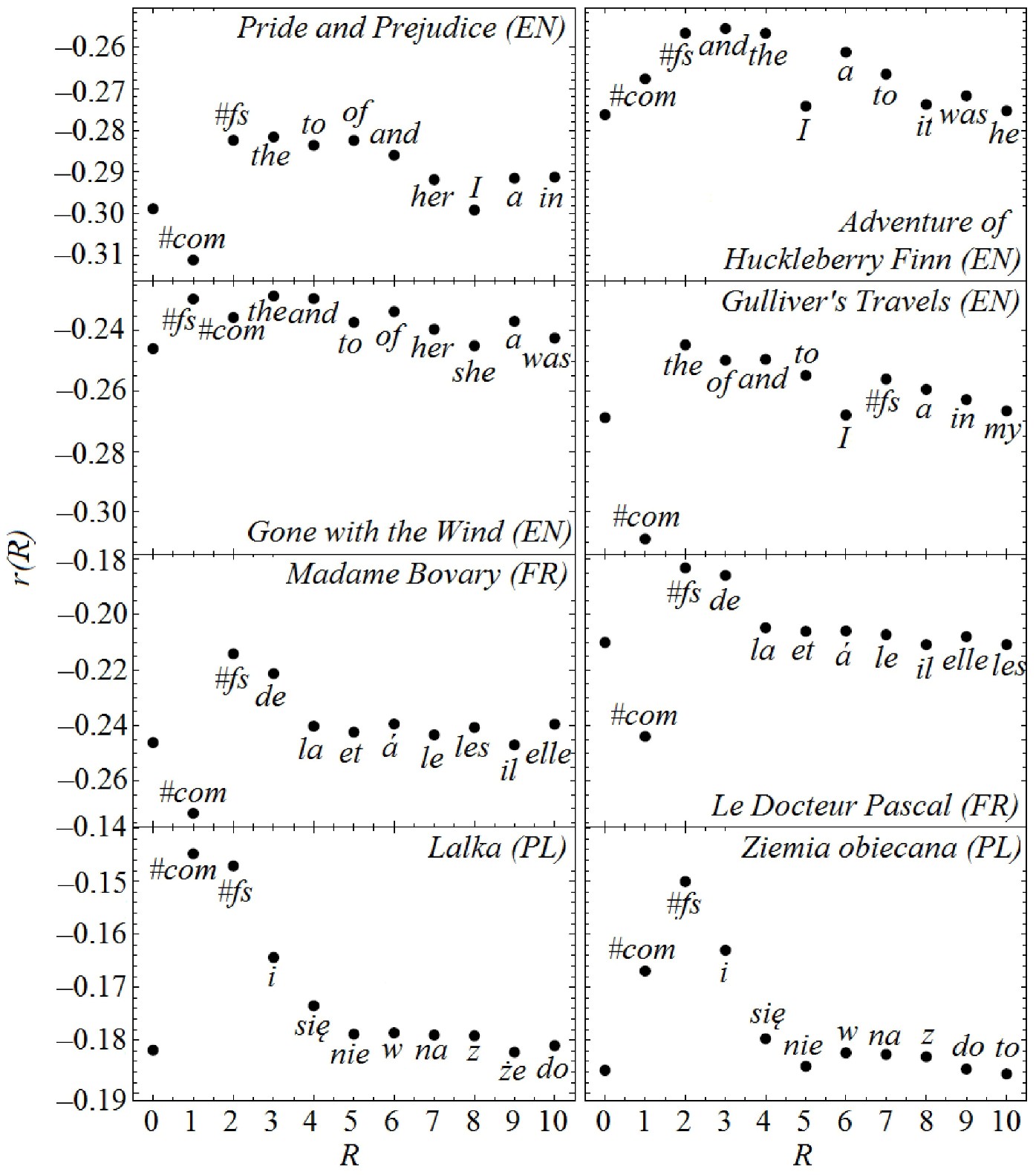}
\caption{The global assortativity index $r(R)$ for the networks representing different novels. For each novel, the rank $R=0$ denotes the complete network with all $n$ nodes, while the lower ranks $1 \le R \le 10$ denote the incomplete networks with $n-1$ nodes obtained by removing a node corresponding to the word ranked $R$ in the Zipf distribution for this novel.}
\label{fig::gai.removed}
\end{figure}

Now we consider another property of nodes, i.e., the indicators how important for the network structure their presence is. In other words, we study how the removing of particular nodes can impact the overall network stucture expressed in terms of the global network measures. We look at three such measures: the average shortest path length: $L = \sum_i \ell_i$, the global clustering coefficient: $C = \sum_i C_i$, and the global assortativity coefficient $r$:
\begin{equation}
r = {\sum_{ij} (\delta_{ij} - {k_i k_j \over 2 e}) \over \sum_{ij} (k_i \delta_{ij} - {k_i k_j \over 2 e})},
\end{equation}
where $e$ is the number of edges in the network and $\delta_{ij}$ equals 1 if there is an edge between the nodes $i$ and $j$ or 0 otherwise. Due to the same reason as before, we first calculate the corresponding quantities $L^{\rm R}$, $C^{\rm R}$, and $r^{\rm R}$ for the randomized text samples (100 independent realizations) and consider them the reference values determined solely by the frequencies of particular items and by neither grammar nor context. We thus consider the values of $L(R)/L^{\rm R}(R)$ (Fig.~\ref{fig::aspl.removed}), $C(R)/C^{\rm R}(R)$ (Fig.~\ref{fig::gcc.removed}), and $r(R)/r^{\rm R}(R)$ (Fig.~\ref{fig::gai.removed}) and expect them to be related to grammar and context largely. For different text samples (novels), we compare the corresponding values calculated for a complete network with all the nodes present (denoted by the abscissa $R=0$) and for 10 incomplete networks obtained by removing a given highly ranked node according to the Zipf distribution ($1 \le R \le 10$).

\begin{figure}[!h]
\centering
\includegraphics[scale=0.88]{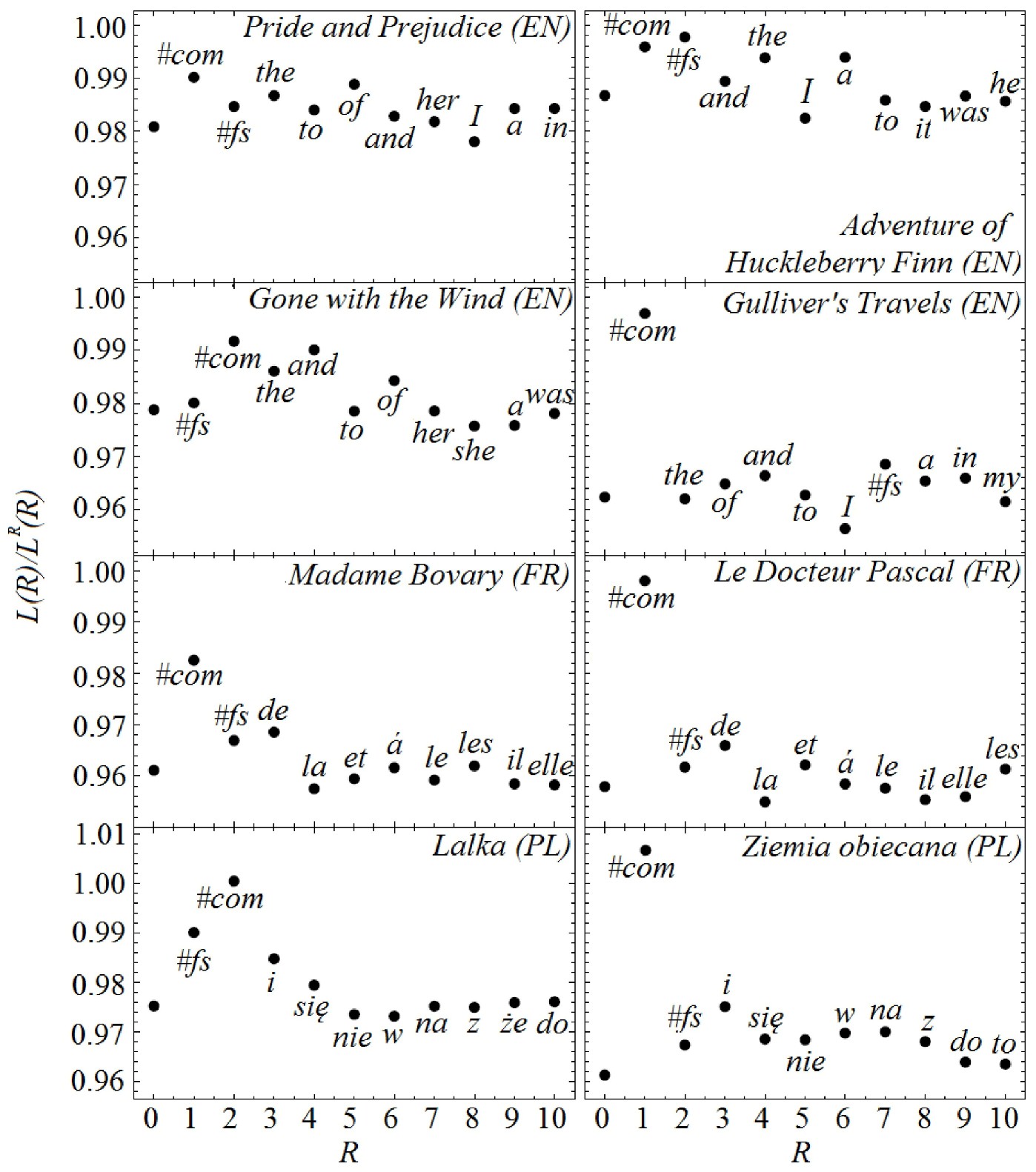}
\caption{The average shortest path length $L(R)$ for the networks representing different novels (the same as in Fig.~\ref{fig::aspl.removed}) divided by its counterpart $L^{\rm R}(R)$ calculated for the null model of random texts with the same Zipf distribution.}
\label{fig::aspl.removed.random}
\end{figure}

\begin{figure}[!h]
\centering
\includegraphics[scale=0.88]{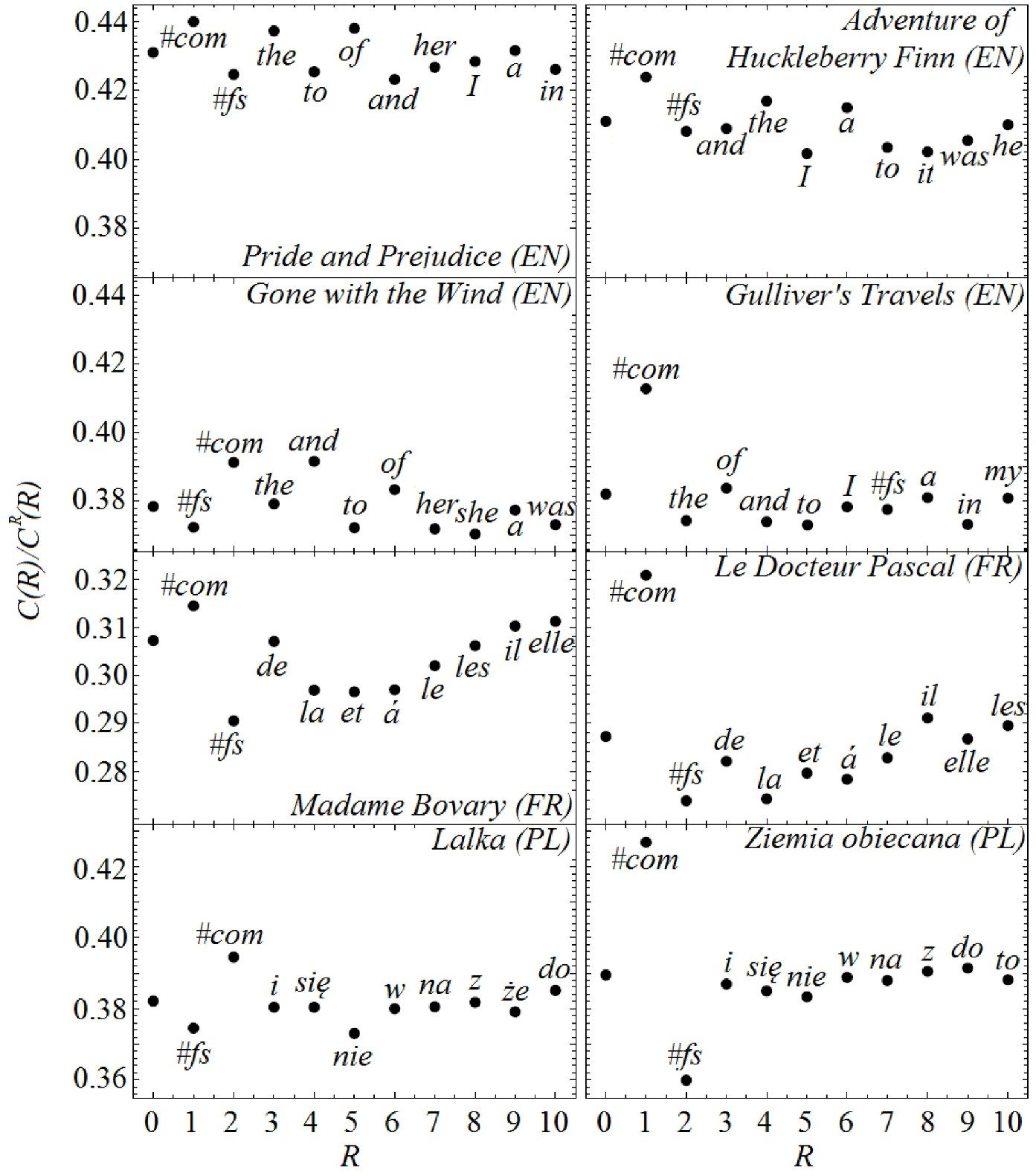}
\caption{The global clustering coefficient $C(R)$ for the networks representing different novels (the same as in Fig.~\ref{fig::gcc.removed}) divided by its counterpart $C^{\rm R}(R)$ calculated for the null model of random texts with the same Zipf distribution.}
\label{fig::gcc.removed.random}
\end{figure}

\begin{figure}[!h]
\centering
\includegraphics[scale=0.66]{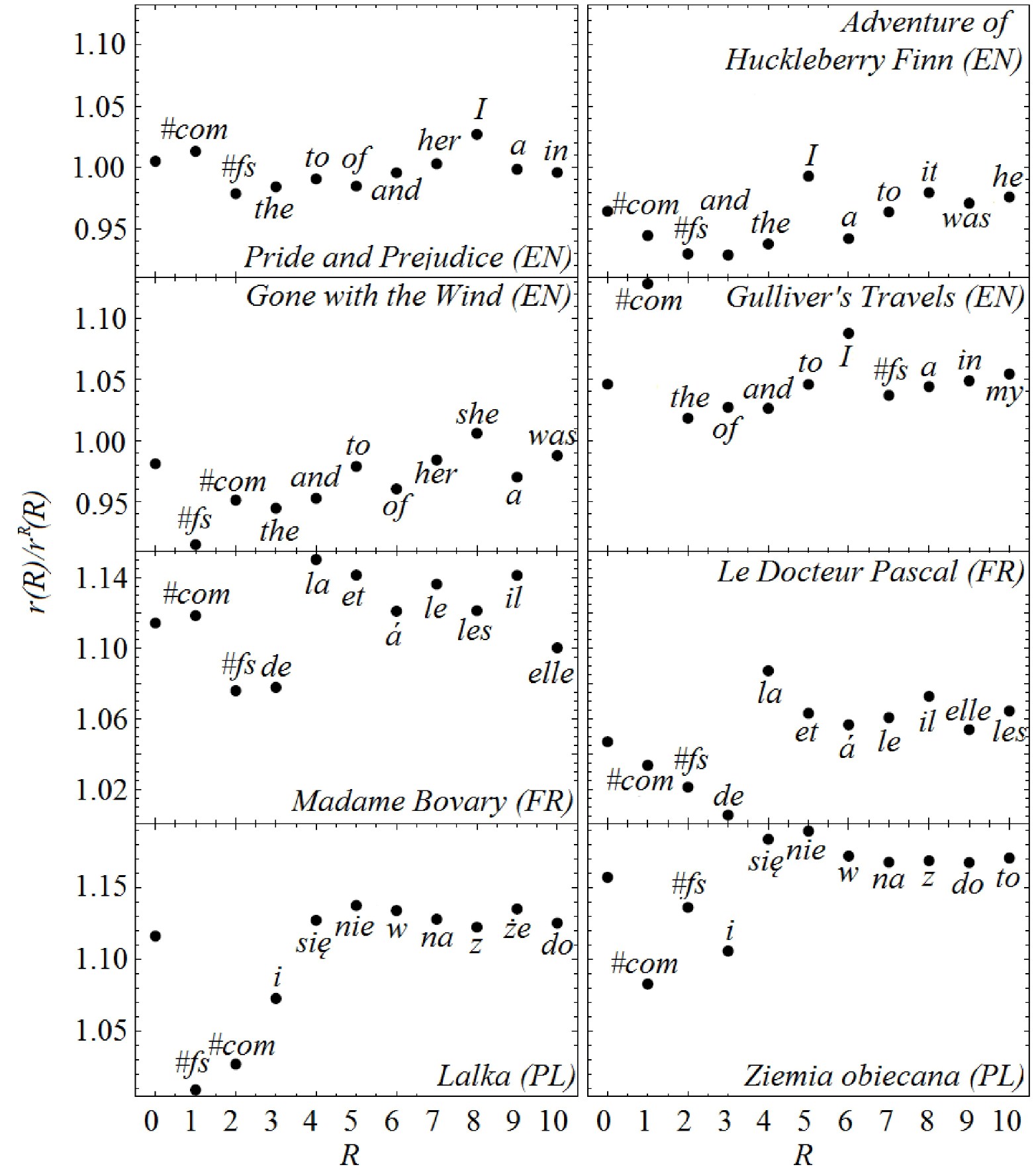}
\caption{The global assortativity index $r(R)$ for the networks representing different novels divided by its counterpart $r^{\rm R}(R)$ calculated for the null model of random texts with the same Zipf distribution.}
\label{fig::gai.removed.random}
\end{figure}

In each case, by removing one of the highly connected nodes, ASPL becomes longer than for the complete network and this is not surprising since the network loses one of its hubs. This increase of $L(R)$ is different for different ranks and different novels $-$ see Fig.~\ref{fig::aspl.removed}, but a rule is that, statistically, the lower the rank (the larger $R$) is, the smaller is the change in $L(R)$ (for a particular novel there might be some exceptions). This rule comes from the fact that in the word-adjacency networks removing a strong hub is more destructive for the network than removing some less connected node. This means that in a typical situation $L(R)$ alters its value the most for comma and the full stop since they occupy the highest ranks, while the observed changes for the function words are smaller. This picture substantially changes if we look at the rescaled quantity: $\lambda(R) = L(R)/L^{\rm R}(R)$ that is free of an item's frequency contribution to ASPL. Fig.~\ref{fig::aspl.removed.random} shows that, except for \#com, $\lambda(R)$ does not exhibit any significant dependence on $R$ and excluding a particular node does not alter its value much as compared to the complete network. Typically, the rescaled ASPL is restricted to a narrow range of $0.95 < \lambda(R) < 1.0$ and this means that the correlations present in the text samples shorten effectively the paths between the nodes as compared to the random network, but this is a small effect. The case of comma is slightly different as, for some texts, the network without this node shows $\lambda \approx 1.0$, i.e., the residual network has the same $L(R)$ as the random one. Presence of this property of comma is text-dependent and it does not seem to be a property of written language. Moreover, it should be stressed that, even if one considers comma, the range of the $\lambda(R)$ variability for different nodes is small.

The global clustering coefficient $C(R)$ and the assortativity index $r(R)$ present a more variable behaviour after removing a hub as these quantities can either increase, remain stable, or decrease. This behaviour obviously depends on a contribution of each particular node to $C$ and $r$ for $R=0$. For the clustering coefficient, a statistical rule is that without particular nodes $C(R)$ does not differ much from its complete-network counterparts. Only for the node representing comma, $C(R)$ can increase more significantly and the network becomes more clustered (Fig.~\ref{fig::gcc.removed}). This can partially be explained by an observation that in all the considered languages commas can mediate words whose direct neighbourhood is unlikely due to rules of grammar. Since $C(R)$ depends on an item's frequency, in Fig.~\ref{fig::gcc.removed.random} we show the rescaled coefficient: $\kappa(R) = C(R)/C^{\rm R}(R)$ whose values are related to the random model. Now the networks without \#com and the ones without other nodes show comparable values of $\kappa$ with only small difference (up to 10\%) for some texts.

As regards the assortativity index $r(R)$, the majority of hubs in the word-adjacency networks (like, e.g., \#fs, articles, and the most frequent conjunctions) can be considered disassortative separators, so after their removal, the overall assortativity index increases (Fig.~\ref{fig::gai.removed}). Of course, since this is only a statistical observation, particular cases may show different behaviour like, e.g., comma, which sometimes acts like a disassortative separator and sometimes like an assortative one. In order to remove the approximately monotonuous dependence of $r$ on rank $R$, we look at the rescaled assortativity index: $\rho(R) = r(R)/r^{\rm R}(R)$. We see that now this dependence is absent and that both the punctuation marks and the function words exhibit comparable values of $\rho$ (see the narrow range of the vertical axis in each panel of Fig.~\ref{fig::gai.removed.random}).

\section{Conclusion}

Punctuation marks are among the most common objects in written language. They do not play purely grammatical roles, but they also carry some semantic load, similar to such words like articles, conjunctions, and prepositions. This opens space for putting a question whether the punctuation marks may be included in any lexical analysis on par with the ordinary words. In this work we addressed this question by comparing the statistical properties of the most common punctuation marks and words using two approaches. We observed that the punctuation marks locate themselves exactly on or in a close vicinity of the power-law Zipfian regime as if they were ordinary words. Moreover, their inclusion acts towards restoring of the Zipf power-law from the more flat Zipf-Mandelbrot behaviour. We drew the same conclusion from an analysis of the word-adjacency networks, in which words, full stops (the aggregated sentence-ending punctuation marks), and commas were considered nodes. In such networks, the punctuation marks are more important than typical nodes: they play a role of the hubs (together with the most frequent words). Despite some minor, quantitative-only differences, topology of such networks and their growth is similar from the perspective of punctuation marks and from the perspective of words. Quantitatively, it is expressed by the node-specific average shortest path length, the local clustering coefficient, the local assortativity, and their global counterparts. These results are qualitatively invariant under language change even for the languages belonging to different Indo-European groups. Regarding the quantitative viewpoint, we do observe certain systematic differences of the network properties between different text samples (including different languages), but considering them here is beyond the scope of this work. A related study will be presented and discussed elsewhere.

By taking all these outcomes into consideration, the principal conclusion from this study is that punctuation marks are almost indistinguishable from other most and medium common words (both the function and the lexical ones) if one investigates their statistical properties. Since the punctuation marks have also non-neglectable meaning, we advocate their inclusion in any type of the word-occurrence and the word-adjacency analysis making it to be more complete. Incorporation of the punctuation marks into an analysis extends its dimensionality and, therefore, it opens more space for possible manifestation of some previously unobserved effects. That this can in fact be fruitful and bring important results, the best example is Ref.~\cite{drozdz2016} where we showed that the sentence length variability can be multifractal for specific (written with the stream-of-consciousness narration) group of texts, while for other texts it remains monofractal. Multifractality is inherently accompanied by burstiness. In the present context this burstiness in the sentence length thus translates itself into analogous effects in the recurrence times (measured by a separation of two consecutive occurrences of the same item) of the full stops. At the same time, however, the recurrence times of the most frequent words appear to be much less bursty as it was also documented in the same Ref.~\cite{drozdz2016}. This latter effect goes in parallel with an observation made earlier~\cite{altmann2009} for the most frequent words. Interestingly, according to the same paper~\cite{altmann2009}, the burstiness is however often observed for the non-function words of high and medium ranks. The related intricacy in fact further supports our thesis that, from the statistical point of view, the punctuation marks are surprisingly similar to the regular words, even though at some angles they resemble more the most frequent function words, while at other angles they resemble more the non-function words. We expect more interesting results will be obtained in future from analyses, in which the punctuation marks are not neglected.

\section{Acknowledgement}

We thank the anonymous referees for very interesting and insightful suggestions that led to significant extensions and improvement of this paper.

\section{Appendix}

The books used in our analysis (asterisks denote the corpora-forming books):

\vspace{0.2cm}

English: George Orwell {\it 1984$^*$}, Mark Twain {\it Adventures of Huckleberry Finn}, Herman Melville {\it Moby Dick$^*$}, Jane Austen {\it Pride and Prejudice$^*$}, James Joyce {\it Ulysses$^*$}, Jonathan Swift {\it Gulliver’s Travels$^*$}, Margaret Mitchell {\it Gone with the Wind}.

German: Friedrich Nietzsche {\it Also sprach Zarathustra$^*$}, Franz Kafka {\it Der Process$^*$}, Heinrich Mann {\it Der Untertan$^*$}, Thomas Mann {\it Der Zauberberg$^*$}, Christiane Vera Felscherinow {\it Wir Kinder vom Bahnhof Zoo$^*$}.

French: Alexandre Dumas {\it Ange Pitou$^*$}, Albert Camus {\it La Peste$^*$}, Émile Zola {\it La Terre$^*$}, {\it Le Docteur Pascal}, Gustave Flaubert {\it Madame Bovary$^*$}, Gaston Leroux {\it Le Fantôme de L'Opéra$^*$}.

Italian: Umberto Eco {\it Il pendolo di Foucault$^*$}, Gabriele d'Annunzio {\it Trionfo della morte$^*$}, Giambattista Bazzoni {\it Falco della Rupe o la guerra di Musso$^*$}, Luigi Capuana {\it Giacinta$^*$}, Tullio Avoledo {\it Le Radici del Cielo$^*$}.

Polish: Gustaw Herling-Grudzi\'nski {\it Inny \'swiat$^*$}, Karol Olgierd Borchardt {\it Znaczy Kapitan$^*$}, Walery {\L}ozi\'nski {\it Zakl\c ety dw\'or$^*$}, Stefan \.Zeromski {\it Przedwio\'snie$^*$}, W{\l}adys{\l}aw Reymont {\it Ziemia obiecana$^*$}, Boles{\l}aw Prus {\it Lalka}, Jerzy Andrzejewski {\it Bramy raju}.

Russian: Lev Tolstoy {\it {\foreignlanguage{russian}{Анна Каренина}} (Anna Karenina)$^*$}, {\it {\foreignlanguage{russian}{Война и мир}} (Voyna i mir)$^*$}, {\it {\foreignlanguage{russian}{Воскресение}} (Voskreseniye)$^*$}, Fyodor Dostoyevsky {\it {\foreignlanguage{russian}{Бесы}} (Besy)$^*$}, {\it {\foreignlanguage{russian}{Братья Карамазовы}} (Brat'ya Karamazovy)$^*$}.

Czech: Bohumil Hrabal {\it Taneční hodiny pro starší a pokročilé}.

\end{document}